%% file: FlevoVision-v1.tex
\documentclass[final,5p,times,twocolumn]{elsarticle}

\usepackage{graphicx}
\usepackage{amssymb}
\usepackage{caption}
\usepackage{subcaption}
\usepackage{rotating}
\usepackage{float}
\usepackage{acronym}

\usepackage{titlesec}

\titleformat{\paragraph}
{\normalfont\normalsize\itshape}{\theparagraph}{1em}{}
\titlespacing*{\paragraph}
{0pt}{3.25ex plus 1ex minus .2ex}{1.5ex plus .2ex}

\usepackage{lineno}
\usepackage{amsmath}
\usepackage[super]{nth}
\usepackage{longtable}
\usepackage{tocloft}

\usepackage{array}
\newcolumntype{L}{>{\centering\arraybackslash}m{3cm}}

\usepackage[colorlinks=true]{hyperref}

\biboptions{authoryear}

\journal{Submitted to Computers and Electronics in Agriculture}

\input{acrolist}

\begin{document}

\begin{frontmatter}

\title{Monitoring crop phenology with street-level imagery using computer vision}



\author{Rapha\"{e}l d'Andrimont, \texorpdfstring{$^{1,*}$}, Momchil Yordanov  \texorpdfstring{$^{1,*}$}, Laura Martinez-Sanchez \texorpdfstring{$^{1}$}, Marijn van der Velde \texorpdfstring{$^{1}$}}
\address{ $^{1}$\quad European Commission, Joint Research Centre (JRC), Ispra , Italy \\
 $^{*}$ These authors contributed equally to this work.
}


\begin{abstract}
Street-level imagery holds a significant potential to scale-up in-situ data collection. This is enabled by combining the use of  cheap high quality cameras with recent advances in deep learning compute solutions to derive relevant thematic  information. We present a framework to collect and extract crop type and phenological information from street level imagery using computer vision. During the 2018 growing season, high definition pictures were captured with side-looking action cameras in the Flevoland province of the Netherlands. Each month from March to October, a fixed 200-km route was surveyed collecting one picture per second resulting in a total of 400,000 geo-tagged pictures. At 220 specific parcel locations detailed on the spot crop phenology observations were recorded for 17 crop types (including bare soil, green manure, and tulips): bare soil, carrots, green manure, grassland, grass seeds, maize, onion, potato, summer barley,  sugar  beet,  spring  cereals,  spring wheat,  tulips,  vegetables,  winter  barley,  winter cereals and winter wheat. Furthermore, the time span included specific pre-emergence parcel stages, such as differently cultivated bare soil for spring and summer crops as well as post-harvest cultivation practices, e.g. green manuring and catch crops. Classification was done using TensorFlow with a well-known image recognition model, based on transfer learning with convolutional neural network (MobileNet). A hypertuning methodology was developed to obtain the best performing model among 160 models. This best model was applied on an independent inference set discriminating crop type with a Macro F1 score of 88.1\% and main phenological stage at 86.9\% at the parcel level. Potential and caveats of the approach along with practical considerations for implementation and improvement are discussed. The proposed framework speeds up high quality in-situ data collection and suggests avenues for massive data collection via automated classification using computer vision.
\end{abstract}


\end{frontmatter}


\renewcommand{\thetable}{Table \arabic{table}}
\renewcommand{\thefigure}{Fig. \arabic{figure}}
\renewcommand{\figurename}{}
\renewcommand{\tablename}{}

\section{Introduction}

\ac{SLI} provides a human-like view with a detailed perspective on the environment around us. \ac{SLI} has become ubiquitous with the growth of capturing devices, personalized mapping services (e.g. Google Street View), social media platforms (e.g. Mapillary), and visual positioning systems for autonomous vehicles. Socio-economic, political \citep{gebru2017using}, and environmental \citep{seiferling2017green} applications have been developed by extracting semantic information from \ac{SLI} using computer vision algorithms. In combination with overhead imagery, e.g. from satellites or drones, or other image sources (e.g. LiDAR, social media), data integration enables further automated cross-referencing and on-the-fly analysis. \ac{SLI} is relatively cheap and can be rapidly collected, mapped, and visualized. Various applications are built to take advantage of this, e.g. for rapid damage assessments or inventories after disasters \citep{lemoine2013intercomparison}, to check and/or confirm geo-located data previously obtained through e.g. remote sensing \citep{d2018targeted}, or to purposely collect new in-situ data to improve remotely sensed based algorithms.

Smartly surveyed \ac{SLI} holds a large potential to efficiently source significant sets of in-situ data. Roadside sampling is a viable source of training data for cropland mapping  \citep{waldner2019roadside}. In a previous study on grassland monitoring, \ac{SLI} were used to evaluate Sentinel-1 and -2 based markers indicative for crop type and bare soil targeting specific parcels \citep{d2018targeted}. Similarly, field pictures have been used in many studies to obtain in-situ information on vegetation, land cover and use, and landscapes. For instance, in-situ repeat photographic imagery has been applied to evaluate the phenological development of natural ecosystems \citep{sonnentag2012digital,klosterman2014evaluating,keenan2014net,nijland2016imaging}. Crowdshared pictures on Flickr have been harnessed for biodiversity monitoring \citep{stafford2010eu,barve2014discovering,elqadi2017mapping}. Usually, classification of vegetation and crop type or phenology on the imagery relies on human interpretation (e.g. \citet{deus2016google,d2018targeted,hufkens2019monitoring,d2020detecting}).

Applying deep learning for image recognition is massively speeding up this process. Deep learning is a machine learning method providing a hierarchical representation of a data set by means of various convolutions \citep{lecun1995convolutional}. This method has gained popularity for image processing and data analysis. This is primarily because of its feature learning ability with the automatic extraction of features from raw data - with features from higher levels of hierarchy being formed by the composition of lower level features \citep{lecun2015deep}. The large learning capacity of deep learning algorithms makes them ideal to perform classification and prediction tasks for a variety of  highly complex data \citep{pan2010survey}. Different types of architectures were developed for deep learning networks. \ac{CNN} are most commonly applied to analyze visual imagery. These are designed to recognize visual patterns directly from pixel images with minimal pre-processing, recognizing patterns with extreme variability and with remarkable robustness to distortions and simple geometric transformations \citep{lecun2015lenet}.

Deep learning on imagery has several applications in the agricultural domain \citep{kamilaris2018deep}, including automated robotic weeding \citep{champ2020instance}, crop disease recognition \citep{mohanty2016using}, paddy crop stress detection \citep{anami2020deep}, phenotyping (\citet{namin2018deep}), and precision agriculture \citep{zheng2019cropdeep}. Other relevant applications include automatic information extraction from phenocam network pictures \citep{cao2021identifying}.  \citet{yalcin2017plant} used a \ac{CNN} to classify five different crop types on images captured by field cameras from the Turkish agriculture monitoring network. \citet{hufkens2019monitoring} applied image processing techniques on smartphone camera photographs to derive crop phenological parameters. At the same time, large amounts of plant species occurrence data are being collected with citizen science applications such as Pl@ntNet \citep{goeau2018deep} that provide plant species recognition capability.

While promising, only a few studies have used \ac{SLI} in combination with deep learning to recognize crops. \citet{paliyam2021street2sat} proposed a framework for obtaining geo-referenced crop type labels obtained from
vehicle-mounted cameras and tested in Kenya for maize and of sugarcane. \citet{wu2021identification} collected a specific \ac{SLI} dataset for crops, and evaluated several deep \ac{CNN}s for the classification task. \citet{yan2021exploring} explored Google Street View with deep learning for crop type mapping, indicating that such methods can be efficient and cost-effective. Despite these previous studies, using \ac{SLI} surveying combined with deep learning for crop and phenology capturing has not yet been used at its best potential. 

\subsection{Objectives}
\label{sec:objectives}
In this paper we describe a fully open and operational approach that starts by ground surveying and collecting the \ac{SLI}, proceeds with classifying crops and crop phenological stages using computer vision, and ends by generating parcel level in-situ data. The objectives are the following: 

\begin{itemize}
\item To design and test a cheap approach to efficiently collect high quality ground truth for crop types and crop phenological stages
\item To collect and to share a representative set of side-looking \ac{SLI} 
\item To produce a generic deep learning hyper-parameterization computer vision framework tailored to street view imagery for crop type and phenology
\item To evaluate to what extent majority voting at parcel level improves the accuracy of the crop type and phenology classification
\item To publish all code and training data fully open in one coherent framework (see section \ref{sec:code})
\end{itemize}

\section{Materials and Methods}
The Materials and Methods section describes 1) the study area, 2) the field data collection including \ac{SLI} collection, crop and phenology in-situ observations, along with the geo-spatial data-processing to associate the imagery to parcels, 3) the deep learning framework, 4) the accuracy assessment for crops and phenology stage at picture and parcel level and, finally, 5) the code and computing infrastructure.   

\subsection{Study area}
\label{sec:study_area}
\begin{figure*}[!ht]
\centering\includegraphics[width=\linewidth]{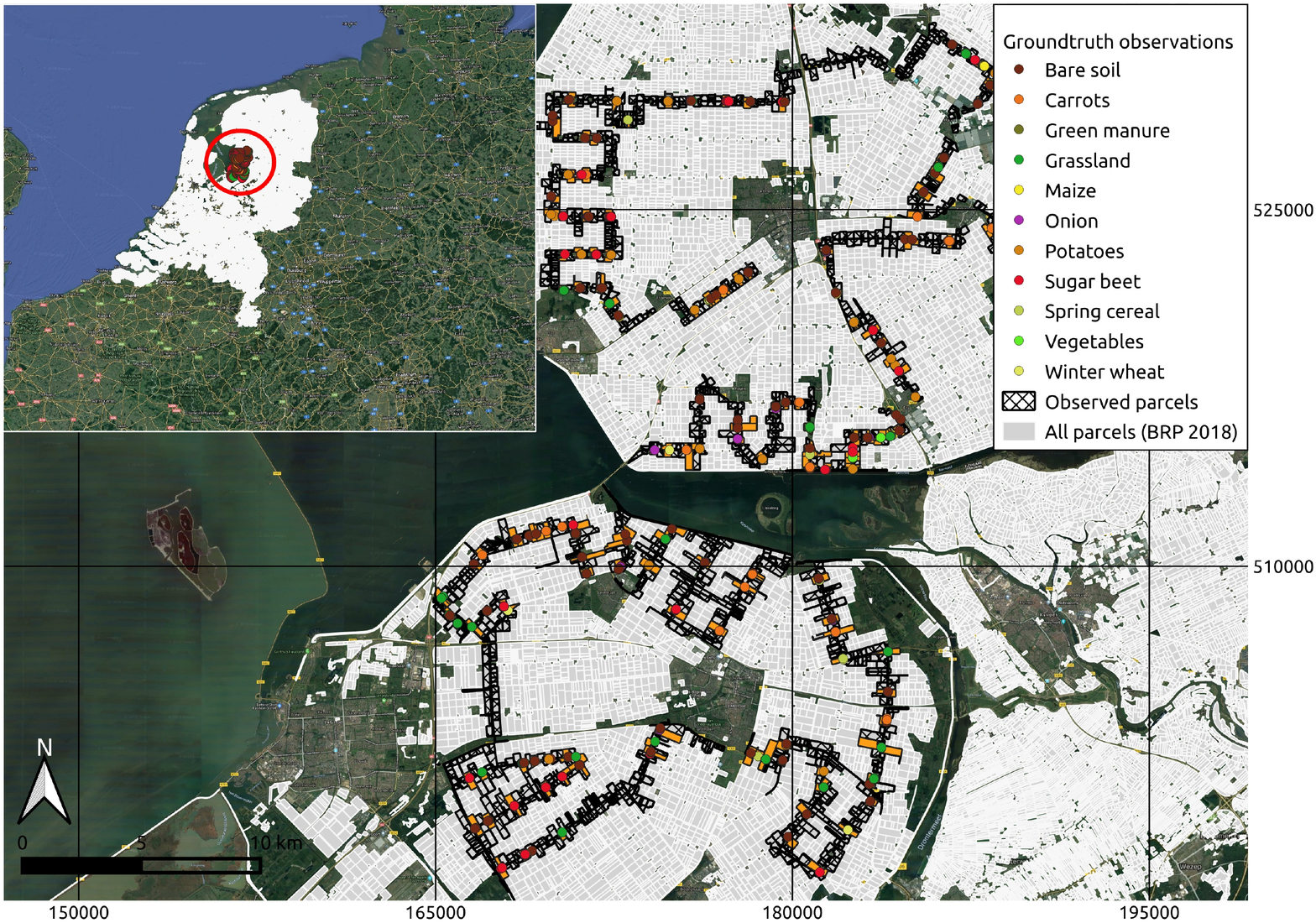}
\caption{The study area is in the province of Flevoland in the Netherlands. The specific study area includes the Oost Flevoland  and Noord-Oost Polder districts. During 8 surveys a total of 1,691 in-situ crop phenology observations were made at 220 parcel locations 
alongside the 200-km surveyed track. Projection of the map: Amersfoort / RD NEW (EPSG:28992).}
\label{fig:Study_site}
\end{figure*}

The study area is located within the Dutch province of Flevoland where agriculture is the dominant land use. The region is one of the most productive ones worldwide with yields close to potential \citep{schaap2011adapting}. Flevoland has ideal features for \ac{SLI} surveying such as a good road network, a large parcel density and crop type heterogeneity. The extent of the study area is shown in \ref{fig:Study_site}, along with the roads surveyed, the parcels and the locations of the collected ground-truth observations. The route followed about 200 km of predominantly inner roads and was surveyed with a 50 km/h speed limit each month during the vegetation growing season. The surveyed track and driving speed were chosen based on the ability to guarantee a relatively consistent image quality, coupled with the ability to stop safely for detailed ground observations. The survey was designed to ensure a statistically meaningful sample density for the main crop types and their phenological dynamics in the area. 

\begin{figure*}[!t]
    \centering\includegraphics[width=\linewidth]{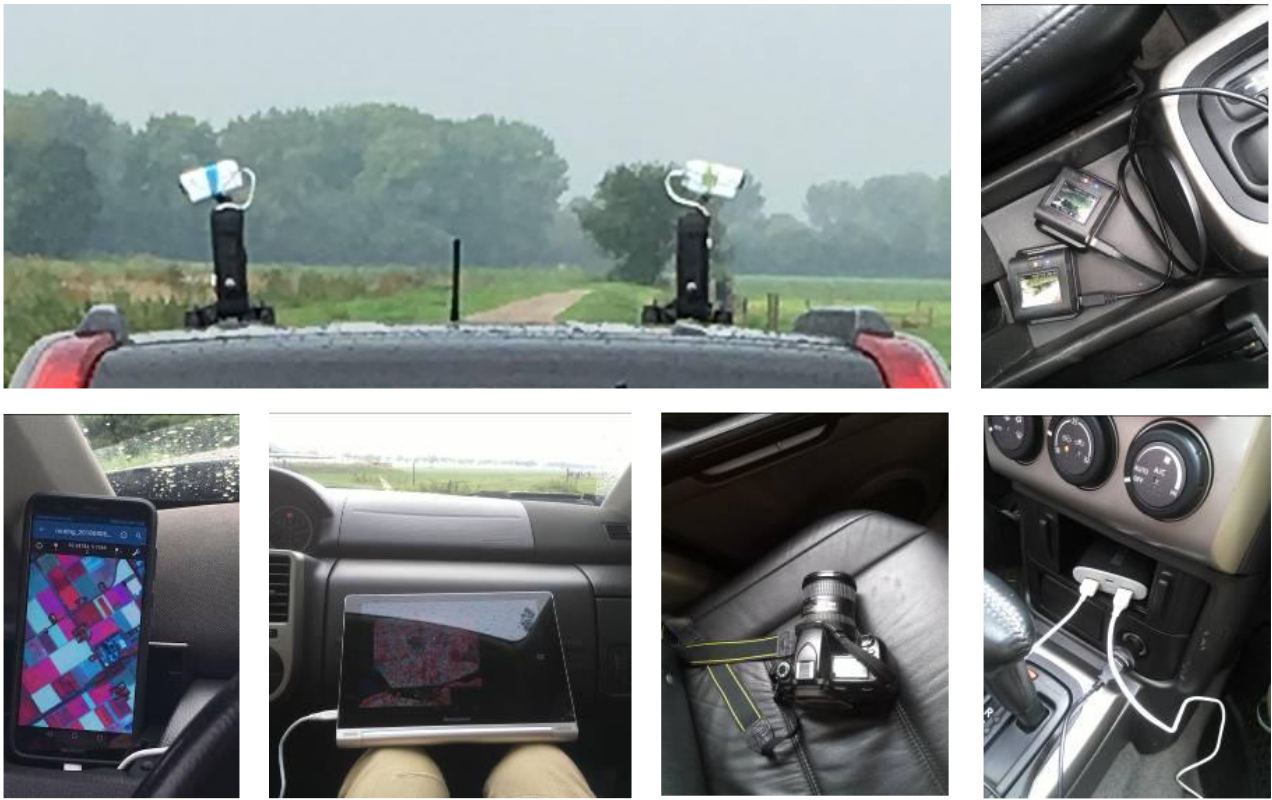}
\caption{The roof camera set-up is composed of two wide-angle side-looking cameras programmed to acquire one picture per second. A high definition camera is used to collect on the spot observations at 220 parcel locations along the surveyed track. The survey routing is supported with actualized Sentinel-2 false-colour imagery that serve as a map base on GPS-enabled mobile phone or tablet.}
\label{fig:camera_set_up}
\end{figure*}

\subsection{Field Data Collection}
A total of eight survey field campaigns were carried out each month from March to October 2018. The surveyors followed the same track during every campaign and collected roughly the same number of pictures - on average around 49,000 \ac{SLI} with GPS location information, and 220 in-situ phenology observations for a total of 17 observed different crop types and their characteristic growth stages.

\subsubsection{Street-level Imagery acquisition}
To collect side-looking SLI, two actions cameras (Sony HDR-AS300R with 128GB miniSD-cards)  were mounted on the roof of the car (Delkin Fat Gekko Camera with a suction cup) as shown in \ref{fig:camera_set_up}. The cameras were equiped with a wide-angle lens (ZEISS Tessar) and were mounted on both (right and left) sides of the roof. The image collection rate was set to one image/second. The lens was chosen based on its ability to capture much of the scene without overly distorting it. The added contrast enhances the difference between the background and the foreground. Two additional reflex cameras (Nikon D80 with Nikon VR 18-200 f 3.5-5.6 G if-ED lenses) were used to collect close-up high-quality detailed images at pre-determined stop locations. Additionally, a Lenovo tablet with PDFmaps was used to store crop type and crop development information. Survey routing and annotation was supported with the use of actualized Sentinel-2 false-colour imagery overlaid on Open Street Map marked with pre-determined stop locations and, when available, the result of the previous survey.

The cameras were set with an identical parameters setting (\ref{tab:sonycamsett}). The images represent what is visible from the side view in both right and left looking directions with about 20\% of the image taken up by the sky and another 20-40\% taken up by road (depending on the distance between the car and the edge of the road). The remaining portion of the image contains the parcel with the relevant crop.

A total of 199,375 \ac{SLI} was collected from the roof-mounted action cameras. A breakdown of the image collection per survey date and image usability (i.e. cropland pictures and pictures for which we have ground truth) is listed in \ref{fig:imagecollection}.

\subsubsection{Phenology Observations}
\label{sec:observations}
For every campaign there were on average 220 ground-truth observations with a minimum sample of 20 parcels per main crop type. The main crops for which ground-truth were collected are: \ac{CAR}, \ac{GMA}, \ac{GRA}, \ac{GRS}, \ac{MAI}, \ac{ONI}, \ac{POT}, \ac{SBA}, \ac{SBT}, \ac{SCR}, \ac{SWH}, \ac{TLP}, \ac{VEG}, \ac{WBA},  \ac{WCR}, \ac{WWH} and \ac{OTH}. For spring and summer crops, the early surveys included the bare soil phase. When surveyors were able to link the bare soil structure to a sown crop class (e.g. ridges for \ac{POT}, seedbed with characteristic row spacing for \ac{SBT} or \ac{CAR}) they were marked as stage 0 for such crop types. For non-specific bare soil stages (e.g. a ploughed or cultivated parcel), the generic \ac{BSO} class was marked with the type of cultivation practice. Similarly, post-harvest observations may include again \ac{BSO} stages or \ac{GMA} if the parcel was re-sown with a catch crop. The sampling points were chosen on basis of geographical spread in the survey area and the possibility to stop safely. At every sampling point, close-up and overview pictures were taken with the Nikon camera, while simultaneously identifying crop type and phenology stage. Phenological stages were recorded in accordance with the BBCH Monograph Stages \citep{meier1997growth} with some notable additions for \ac{BSO} as explained above (introducing classes and generalizations described in section \ref{sec:trainingset} and description of different stages in \ref{tab:allbbch_explained}). 

\subsubsection{Linking SLI to parcels}
\label{sec:in-situ_sample}
\begin{figure*}[!ht]
    \centering\includegraphics[width=\linewidth]{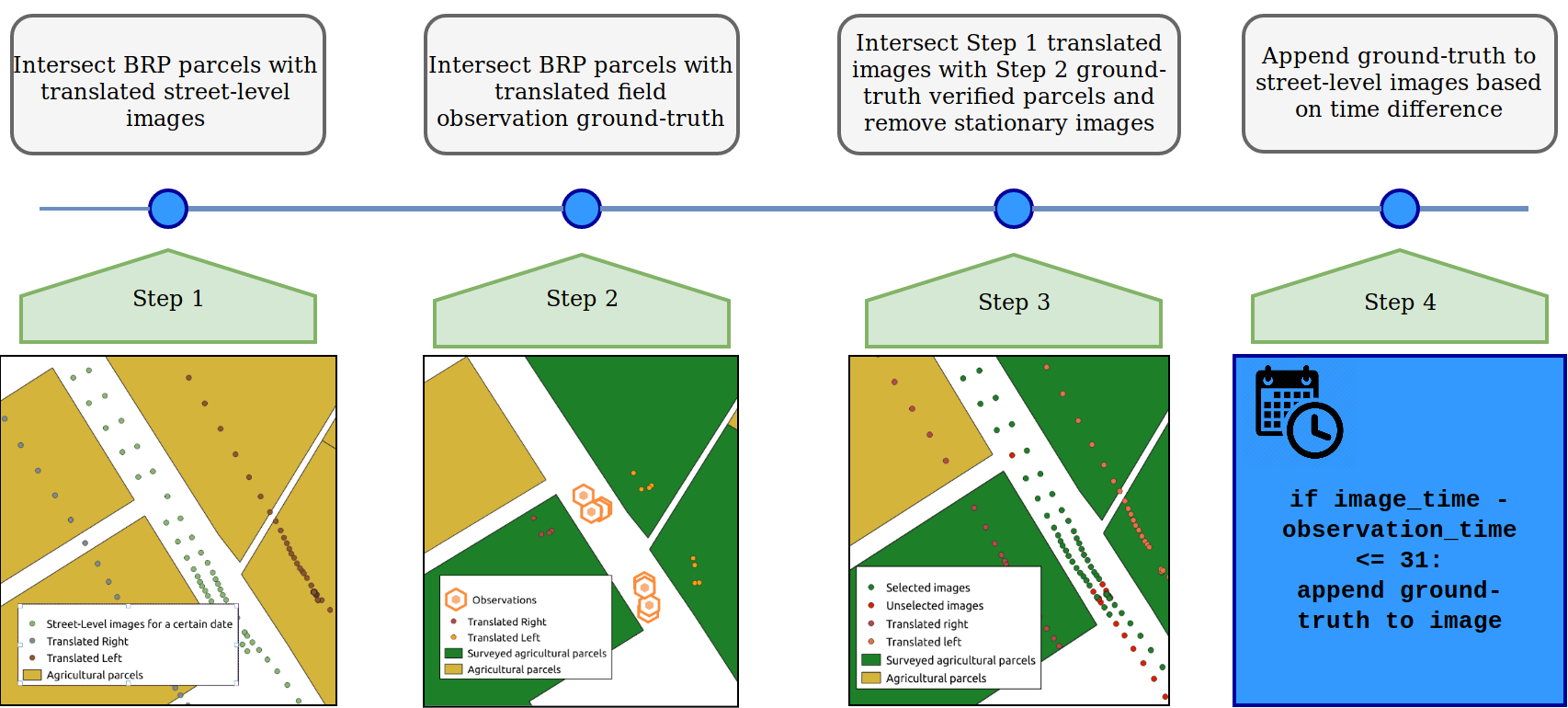}
\caption{Four-step spatial analysis to link observation data and \ac{SLI} located on the road to specific agricultural parcels.}
\label{fig:spAnalysis}
\end{figure*}

After the surveys, each picture and in-situ observation collected is linked to a specific parcel. Agricultural parcel geometries are openly available in the Netherlands as the \ac{BRP} datasets \citep{BRP} and were downloaded for the study. The in-situ observed data (section \ref{sec:observations}) were processed and attached to the corresponding parcel using a four-step geospatial process (\ref{fig:spAnalysis}). 

\begin{enumerate}
\item The \ac{SLI} captured locations (i.e. the GPS measurements on the road from which the pictures were taken) are first transposed 30 meters perpendicularly to the driving direction to the left or to the right side according to the camera (step 1). The geometrical intersection of the transposed locations with the agricultural parcel geometry allows linking most of the pictures to their corresponding parcel.
\item Similarly, the in-situ observations are transposed and linked to their corresponding parcels (step 2) .
\item The parcels with in-situ observation were then selected and duplicate \ac{SLI} at stop locations are removed (step 3). 
\item Finally, the in-situ observation is appended to \ac{SLI} on the basis of the date of the difference in time between when the picture acquisition and the in-situ observations (step 4).
\end{enumerate}

\subsection{Deep Learning}
A deep learning framework was designed and implemented to classify the pictures in a given crop type and phenology class. In this section we describe 1) the choice of network, 2) the selection of training data, and 3) the training and hypertuning of the model.

\subsubsection{Network selection}
Instead of designing a deep learning network from scratch, we decided to use an out-of-the-box model and retrain only the final layer via transfer learning. Transfer learning is the process of taking a model trained for one task, where data is more readily available, and applying it to a new but similar task \citep{pan2009survey}. The study makes use of a network, trained on the large ImageNet data set \citep{deng2009imagenet}, while bootstrapping a new fully-connected top layer to recognize other classes of images. In this sense the approach can be categorized as a network-based, multi-class deep transfer learning for feature extraction (\citet{li2017learning}), applied over an imbalanced test set (i.e. the classes have different number of pictures). Other than reducing data dependence, reducing time to develop the network, and the computational cost of training, this technique has been shown to advance performance in terms of a higher start, higher slope and a higher asymptote \citep{olivas2009handbook}.

The selected \ac{CNN}  comes from the  MobileNet \citep{howard2017mobilenets} family - a set of  open-source, on-device, mobile-first computer vision models for TensorFlow. MobileNets are revolutionary in terms of the simplicity and efficiency of the network, achieving high accuracy results, while keeping the total number of parameters low. MobileNets use depthwise separable convolutions \citep{howard2017mobilenets} instead of regular convolution reducing the number of parameters compared to the regular network with the same depth in the nets. This results in lightweight deep neural networks. The MobileNet version 2 used here represents an adequate trade off between accuracy and computing requirements as they include the residual connections and the expand/projection layers in their architecture \citep{sandler2018mobilenetv2}.

\subsubsection{Training set selection}
\label{sec:trainingset}

Following preliminary tests, we selected 400 images per class for training. To limit spatial auto-correlation in the sample, each parcel was sampled by a single image iteratively until the number of 400 images per class was reached. This process ensures that no parcel gets over-sampled due to its relatively large size (and consequently a larger number of images), compared to the other parcels of the same class. To create the training data set three steps were followed, 1) classes not separable visually were combined, 2) pictures taken at the edge of parcels were removed as they could contain two different crops and 3) pictures were visually inspected and re-labelled as 'other' when the crops was not visible. 

Classes originally distinguished in the field were combined to avoid confusion as described in \ref{tab:generalizations}. Firstly, in the pre-emergence stage (i.e. BBCH stage is 0) of some crops, it is impossible to observe the canopy elements that would provide the basis for class distinction (however, sometimes the soil structure may contain clues); secondly when the difference between stages is minimal and not observable on SLI, and finally where the ripening and senescence phases are too similar to be distinguished. Importantly, these aggregations were done after inferencing with the trained model multiple times, examining the confusion matrices, and applying common sense as to the level of detail that is discernible on a re-sized full resolution picture (see sec. \ref{S:limitations}).

A specific metric used to identify pictures close to the edge of the parcel was calculated on the basis of the distance between each \ac{SLI} and the centroid of the corresponding parcel. This results in a ratio ranging between one (picture is further from the centroid) to > 1 (the picture is closest to the centroid). For most of the parcels, this ratio could be used to mark pictures potentially covering two parcels.
When selecting images for the training set it is important to exclude those pictures, because they do not provide a good view on the crop. A value of 0.95 proved satisfactory.

In the third step pictures were visually inspected and re-labelled as 'other' when the crops was not visible. The 'other' class includes all the images that have a noticeable visual obstruction between the camera and the parcel. 
These can be in the form of dense vegetation, embankments, automobiles, trucks, or various stationary objects that are placed adjacent to the road that prevent a direct view on the parcel. The 'other' class was manually selected and is included in the classification task, meaning it is one of the classes the network is trained on. Although there are many images with such visual obstructions present in the inference set, the study uses the 'other' class as a means of detecting, flagging, and removing such images before the classification task.

\begin{table*}
\footnotesize
\centering
\begin{tabular}{LLL}\hline
\textbf{New class }& \textbf{Old class} & \multicolumn{1}{m{6cm}}{\textbf{Description and rationale}}\\
\\\hline

BSO0 & BSO1, BSO3, BSO4, BSO5, BSO6, ONI0, POT0, SBT0, WWH0 &  \multicolumn{1}{m{6cm}}{Although \ac{SLI} has a potential to separate specific bare soil structures, e.g. by cultivation type, our current study focuses on vegetative crop stages.}\\\hline
  GMA2 & BSO2 & \multicolumn{1}{m{6cm}}{Late season bare soil stage 2 is erroneously described by surveyors as Green manure.}\\\hline
  GRA1 & GRA2, GRA3 & \multicolumn{1}{m{6cm}}{One class for all post-emergence grass BBCH stages due to visual similarity.}\\\hline
  ONI4 & ONI41 & \multicolumn{1}{m{6cm}}{Essentially the same class with minor difference, visible only upon very close-up inspection.}\\\hline
  ONI48 & ONI45 & \multicolumn{1}{m{6cm}}{From 50\% to 80\% of expected bulb/shaft diameter visible. Detail not visible on such images.}\\\hline
  SBT1 & SBT11 & \multicolumn{1}{m{6cm}}{Essentially the same class with minor difference, visible only upon very close-up inspection.}\\\hline
  WWH8 & WWH9 & \multicolumn{1}{m{6cm}}{Ripening and senescence phases for winter wheat look very similar. Detail not visible on such images.}\\
  \hline
\end{tabular}
\caption{Class generalizations.}
\label{tab:generalizations}
\end{table*}


\subsubsection{Training and hyper-tuning}
\begin{figure*}[!ht]
\centering\includegraphics[width=\linewidth]{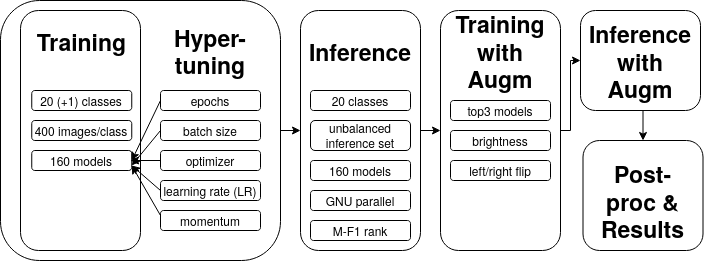}
\caption{Deep learning processing framework. \textit{Hyper-tuning} lists the parameters that the network is optimizing, \textit{Training} relates to the training set selection ('+1' explained in Section \ref{sec:trainingset}) and training of the models, \textit{Inference} shows the unbalanced inference set (explained in Section \ref{sec:validation}), and \textit{Training with Augm} lists the augmentations applied for the second round of training over the best performing models, where they are selected based on the best from each performance level (see Section \ref{sec:validation}), for \textit{Post-proc \& Results} see \ref{fig:resultsChain}.}
\label{fig:dlprocchain}
\end{figure*}

The MobileNet architecture divides the training set into training, test, and validation bins. As such our original training sample benchmark of 400 images per crop phenological class is divided to 80\% for training, 10\% for testing, and 10\% for validation. A hyper-tuning framework was implemented to find the model parameters (i.e. epochs, batch size, optimizer, learning rate, momentum) achieving the best accuracy. Following this hyper-tuning, the top performing models are then further trained with image augmentations. A schematic representation of the deep learning processing framework can be found in \ref{fig:dlprocchain}. Hyper-tuning a model is the process of systematically trying various parameter settings in order to find the best fit for the relevant classification task.

Hyper-tuning was performed with the following parameters: type of MobileNet, number of training samples, batch size, optimizer, learning rate, learning momentum, and a series of image augmentations. Preliminary bench-marking indicated that the best performance, relative to training time was achieved with the primary model (width multiplier 1, and image size re-sampled to 224x224). For batch size, the best working values were batch sizes of 512 and 1024 images. For optimizers the experiment was ran using Adam and Gradient Descent (GD). A two-step process was applied for learning rate and momentum. First, initiate a random search within a 2D space with a good spread of values within the space. Second, models were trained in each combination to see in which domain of values the model performed best. The hyper-tuning was implemented through a bash script that simultaneously creates the directory tree, generates the files needed to run the network, and performs the relevant combinations of hyper-parameters.

\subsection{Independent inferencing and parcel level classification}
\label{sec:validation}

\begin{figure*}[!ht]
\centering\includegraphics[width=\linewidth]{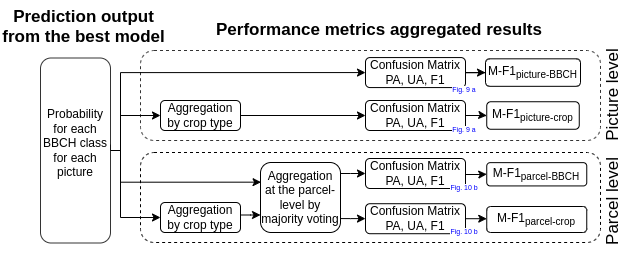}
\caption{The output predictions of the best model provide the probability for each picture to be in the different BBCH classes. Performance metrics are computed for both crop type and phenological stage and predictions are evaluated at picture and parcel level. Finally, the confusion matrices, \ac{PA}, \ac{UA}, \ac{F1} and the \ac{M-F1} results are computed for all performance levels. Resulting figure references are in blue.}
\label{fig:resultsChain}
\end{figure*}

A total of 160 models were trained in parallel to identify the best combination of parameter values. After this first round of training, the top-three model configurations are identified based on the \ac{M-F1}  achieved during the independent inferencing. Then, a second round of training takes place on these three models including image augmentations to boost the accuracy of the best models. Finally, the best model is selected based again on a \ac{M-F1} ranking. A schematic representation of the processing steps for calculating accuracy is presented in \ref{fig:resultsChain}.

A stand-alone inferencing with the best model was also performed on a large set of SLI. This allows us to evaluate the model's performance independently from the training, test, and validation cycle in a real-life configuration. The classification accuracy is also evaluated when using majority voting at parcel level. Since multiple SLIs are taken at each parcel, at each time, an increased confidence in classification accuracy would improve the operational robustness of the  methodology to collect in-situ data for cropped parcels.

Prior to calculating performance metrics on any level (Sections \ref{sec:bbchlevel}, \ref{sec:clevel}) there is a reduction of the inference set. As noted in section \ref{sec:trainingset}, the introduction of an 'other' class is done in order to flag images without crop. To be able to cross compare model performance, it is necessary to maintain the same number of images. Therefore all images labelled as 'other' by any of the models, are identified and subtracted from the inference set prior to any accuracy calculation. The same was done for the bare soil class ('BSO0').

\subsubsection{Performance metrics}
\label{sec:metrics}

The presented study can be categorized as a multi-class imbalanced classification task, where all the classes matter equally \citep{sun2009classification}. Because standard performance metrics, such as overall accuracy and error, have been shown to be misleading in such cases \citep{branco2015survey}, the main metric, by which a comparison of model configurations is carried out, is the \ac{M-F1}. The M-F1 gets calculated from a confusion matrix using the \texttt{confusionMatrix} function from R's \texttt{caret} package. It outputs information on, among others, the following variables: Precision, Recall, \ac{F1}.  Precision (\ac{UA}) is a measure of how many of the predicted positive outputs are actually positive, while recall (\ac{PA} or sensitivity) is a measure of how many of the actual positives the model captured by labelling it as positive are so, and \ac{F1} is a balanced metric between the two. Because \ac{F1} is a score given to each class, the paper makes use of an aggregation of every per-class value into a single number, \ac{M-F1}. \ac{M-F1} is calculated by taking the mean of all \ac{F1} scores, rather than taking the mean of the precision and recall and then calculating \ac{M-F1}, in accordance with the literature (\citet{opitz2019macro}).

\subsubsection{Performance levels and parcel aggregation}
\label{sec:plevels}

Performance metrics for computer vision classification tasks are usually calculated on a per-image basis. This means that for each image the network outputs a vector of probabilities with the likelihood for each class. This vector must sum to 1. For the 160 models, the \ac{M-F1} of each model is calculated by taking the mean of the \ac{F1} scores for the inference set for each class. Four performance levels are considered in this study. At the picture level (i.e. considering each image from the inference set), the \ac{M-F1} is computed for the phenological stage ($M-F1_{picture-BBCH}$) and for the crop type level by aggregating the pictures of the same crop type in the same class ($M-F1_{picture-crop}$). In addition to these two metrics,  similar metrics are computed with an aggregation at the parcel level using a majority voting approach resulting in a \ac{M-F1} metric at the parcel for both phenological stage ($M-F1_{parcelBBCH}$) and crop type ($M-F1_{parcel-crop}$).

\subsection{Code and Computational infrastructure}
\label{sec:code}

All the code developed for this study is available openly on the following repository: \url{https://github.com/Momut1/flevoland}. The working environment was set up as a docker image.  The processing pipeline is fully automated for ease of reproducibility. It is fitted to work both locally and in a cloud through the use of a docker environment. Nearly the entire pipeline is made to function through the calling of dedicated shell scripts that separately accomplish the hyper-tuning, inferencing, and results calculation. For more information consult the \textit{readMe} of the git repository. Part of the processing was done on the JRC \ac{BDAP}, an in-house, cloud-based, versatile, petabyte-scale platform for heavy-duty processing \citep{soille2018versatile}. The offered GPU services work on a NVidia GeForce GTX 1080 Ti with 11GB memory, CUDA version 10.1, and CUDA driver version 418.67. Pre-processing, launching, and post-processing are done in the JEO-lab layer of the platform in a jupyter notebook docker container, running Tensorflow 1.3.0. Another part of the processing was done on \ac{AWS} \citep{cloud2011amazon} using the EC2 functionality to launch custom virtual machines. The study made use of one virtual machine for training which had the following specifications: G4dn EC2 instance with a NVIDIA T4 GPUs with 32GB memory with an identical CUDA and Cudnn version setting; and one for inferencing, which in turn had the following specifications: Inf1 EC2 instance with 24 CPUs.

\section{Results}
\label{sec:results}

The results are divided into four sections. First, the resulting dataset, along with relevant processing, is presented (Section \ref{sec:datacol}). Second, the best performing models are presented (Section \ref{sec:modelselcetion} ). Thirdly, the ability of the model to discriminate between the respective crop and phenology combination (Section \ref{sec:bbchlevel}) will be shown. Results (\ac{M-F1})  are presented both at image level and parcel level. Finally, the model ability  to distinguish crop types only by aggregating the phenological stages of each crop into a single class is presented in  Section \ref{sec:clevel}. 

\subsection{Data collection}
\label{sec:datacol}

\begin{figure*}[!ht]
\centering\includegraphics[width=\linewidth]{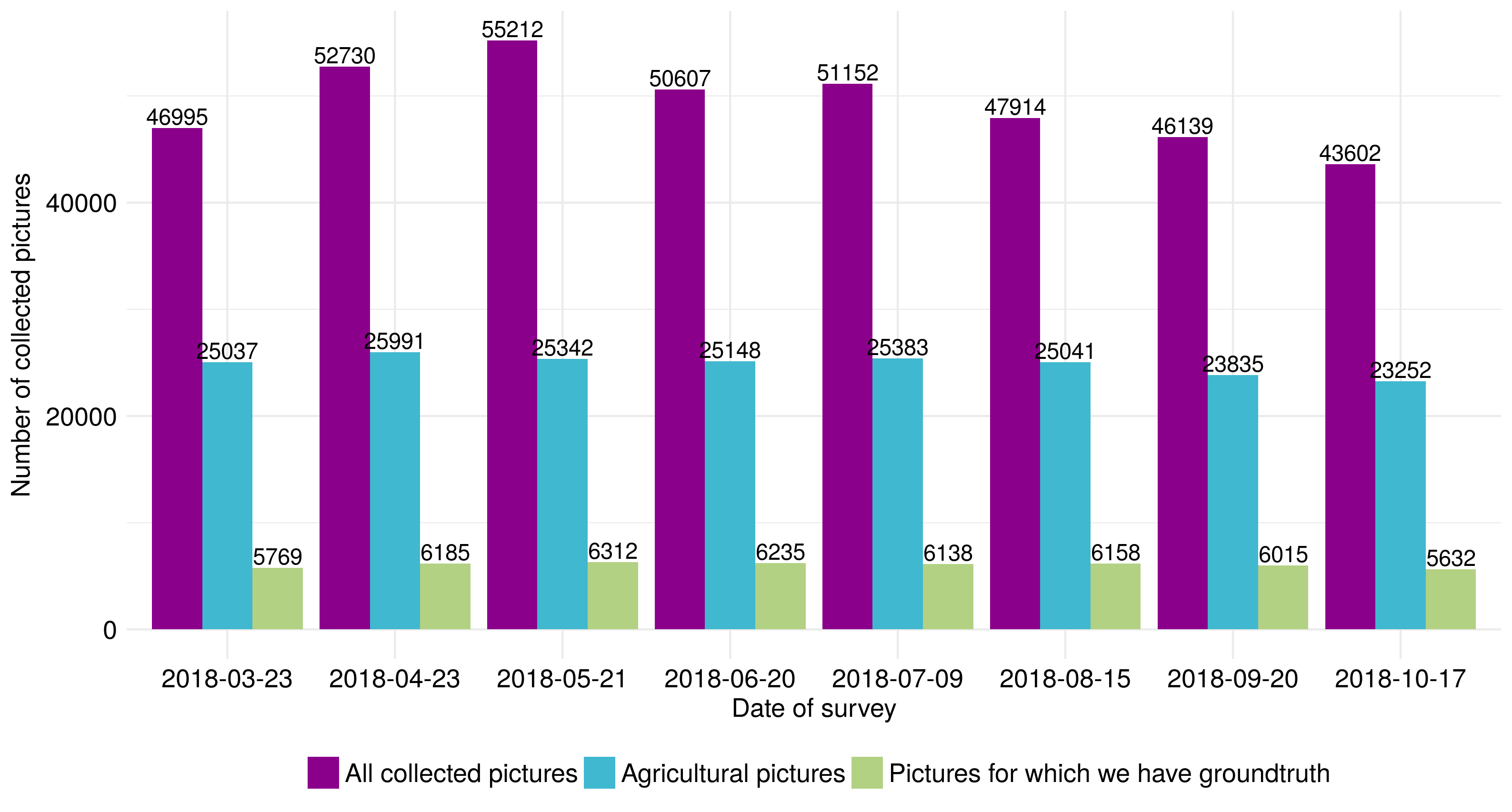}
\caption{Number of image collected of the eight field campaigns from March 2018 to October 2018.}
\label{fig:imagecollection}
\end{figure*}

The collected data is organized in a structured form in a spatial database (PostgreSQL/Postgis) for easy querying and access. Each row represents an image, which is identifiable via a unique file path, also holding a variety of other attributes such as spatial location, capture timestamp, unique parcel ID, crop type according and the BBCH stage from the in-situ observation. A breakdown of image collection for the eight field campaigns can be seen in \ref{fig:imagecollection}. The total number of collected pictures and the pictures of cropped parcels respectively amount to nine and five-fold the pictures with in-situ observations. Photos with in-situ observations (green bar) amount to an overall total of 48,444 and are the result of the geospatial processing described in Section  \ref{sec:in-situ_sample}.

\begin{figure*}[!ht]
\centering\includegraphics[width=\linewidth]{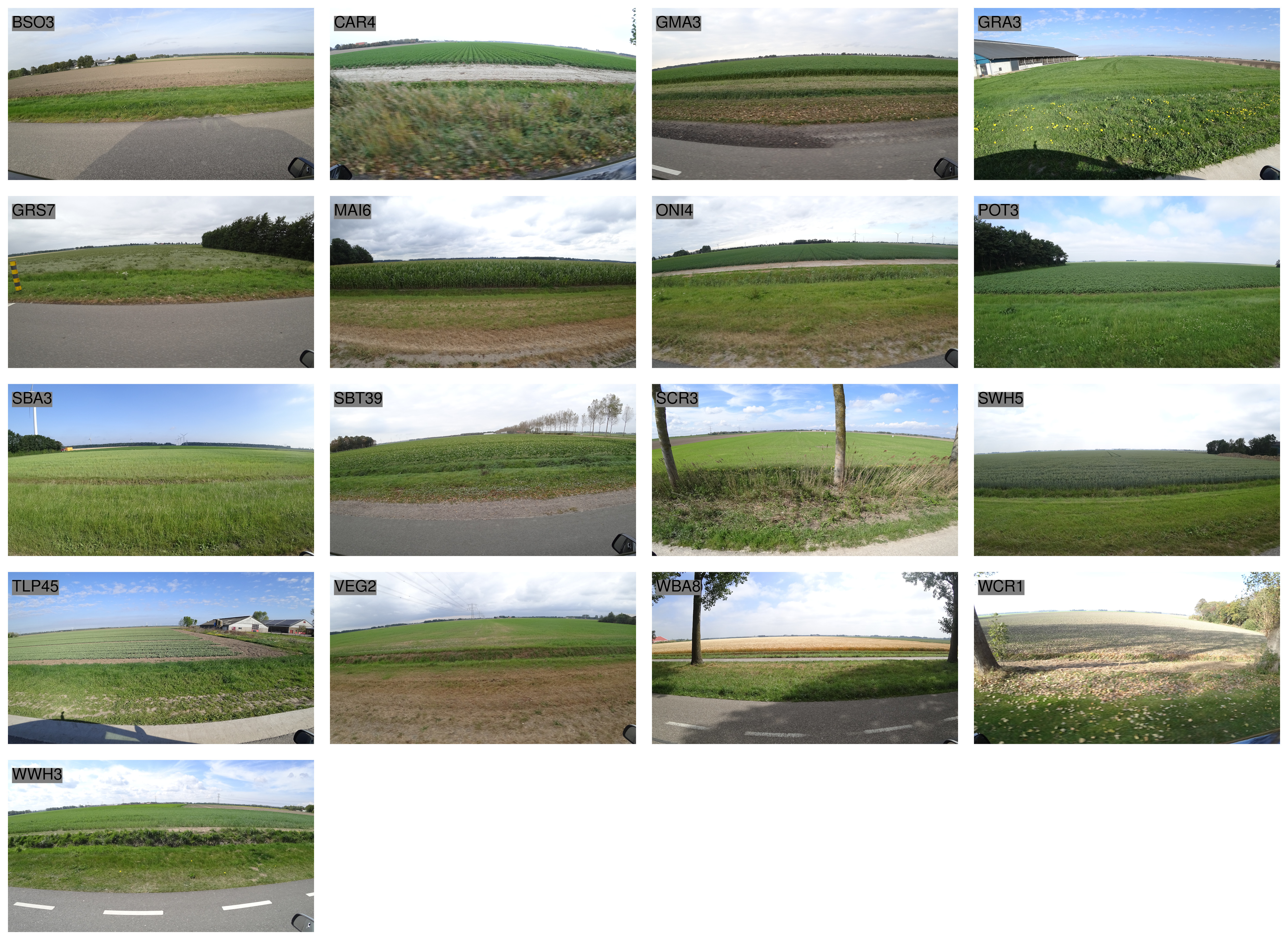}
\caption{Examples of image for each of the 17 different collected crop types for random BBCH stage. In addition to the crop field, the picture could included sky, road, buffer strips, trees, woody features, buildings or other elements of the landscape. The label of the crop along the BBCH is shown on the upper left of the image (see \ref{tab:allbbch_explained} for label details) }
\label{fig:allcrops}
\end{figure*}

An example of random pictures taken for each of the 17 different crops is shown in \ref{fig:allcrops}. These examples of pictures show that the crop parcel of interest covers only a small fraction of the whole scene. Indeed, mainly the sky and the road along with other smaller elements such landscapes features (e.g. trees, hedges, buffer strips) or artificial elements (e.g. building, windmill) could cover a large part. These pictures also illustrate the wide angle lens distortion effect from the cameras allowing to capture an almost 180 degree azimuthal view. 

\begin{sidewaysfigure*}
    \centering
    \includegraphics[scale =0.8]{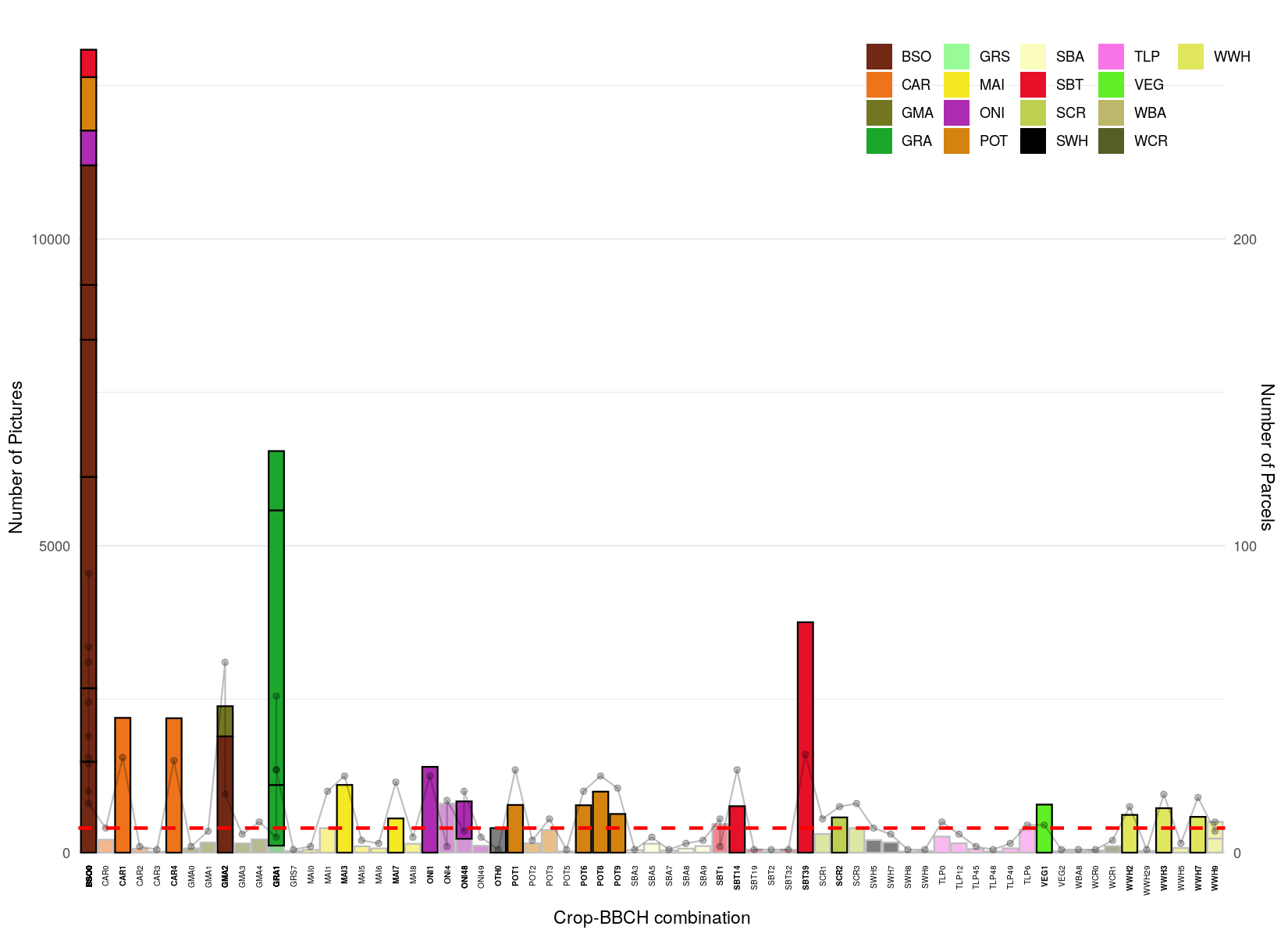}
    \caption{Sample of street-level images for which in-situ information was derived from field observations. The vertical bars represent the number of pictures for each BBCH stage of each crop (see \ref{tab:allbbch_explained} for label details), the connected points represent the number of parcels (on the secondary Y-axis) for each BBCH stage of each crop, and the dotted red line represents the threshold of 400 images. The highlighted bars are the classes that make up the computer-vision experiment. The bars themselves are also stacked to reflect the class generalizations, described in \ref{tab:generalizations}.}
    \label{fig:picsNParcels}
\end{sidewaysfigure*}

 The number of images and parcels per crop and per BBCH stage that make up the complete labelled dataset are shown in \ref{fig:picsNParcels}. The highlighted bars are the classes that make up the computer-vision experiment while the grey ones were discarded as not reaching the 400 pictures required thresholds. The vertical bars represent the number of pictures for each BBCH stage of each crop, the connected points represent the number of parcels (on the secondary Y-axis) for each BBCH stage of each crop. After discarding the classes with less than 400 pictures and  applying the class generalizations from \ref{tab:generalizations}, we obtain a final training dataset of 21 crop/BBCH combination classes which aggregate at crop level to 12 crop classes (including 'other'). Specific classes (i.e. ONI4, SBT1, SCR3, MAI1) with more than 400 images were removed due to distance to centroid and visual screening during training set selection.

The picture dataset is available at \url{https://jeodpp.jrc.ec.europa.eu/ftp/jrc-opendata/DRLL/FlevoVision/training/}.

\subsection{Best model selection}
\label{sec:modelselcetion}
The best model configuration was selected based on the \ac{M-F1} ranking (detailed results can also be found in GitHub \href{https://github.com/Momut1/flevoland/blob/60755e0cd7c15454c83cc81dbae41cc620f9cd65/Results-all-images_R_f1_withParams.csv}{here}). \ref{tab:top3augmented} summarizes the top three performing models; the most accurate is \# 44 with a \ac{M-F1} of \text{59.4\%}, \text{86.9\%}, \text{62.3\%}, \text{88.1\%} respectively for the picture-phenology, parcel-phenology, picture-crop, and parcel-crop (Section \ref{sec:plevels}). The best model configuration (Model \# 44) was built with a GD optimizer, a training batch size of \text{512}, a learning rate of \text{0.08990006381198408}, with augmentations flipping images left and right, training up to the last to final \nth{2999} epoch. The model is also the best in not confusing bare soil with the other classes. This confusion occurs less than 3.2 times compared to both other top models.

\begin{table}[ht]
\centering
\caption{Output for the top three performing models with augmentations. The applied augmentation was a left-right flip. The table shows (in order) the model number, along with the relevant configuration (Learning rate, Batch size, Momentum, Optimizer), the number of images classified as other and bare soil for each model, the training and validation accuracy, and the \ac{M-F1} for each aggregation level.}
\label{tab:top3augmented}
\begin{tabular}{r|lll}
  \hline
\textbf{Ranking} & \textbf{1} & \textbf{2} & \textbf{3} \\ 
  \hline
Model \# &  44 & 107 & 127 \\ 
  Learning Rate & 8.99e-02 & 4.6e-04 & 6.58e-04 \\ 
  Batch Size &  512 & 1024 & 1024 \\ 
  Momentum & 0e+00 & 9.36e-01 & 9.19e-01 \\ 
  Optimizer & GD & Adam & Adam \\ 
    \hline
  Other \# & 668 & 663 & 661 \\ 
  Bare soil \# &  867 & 2166 & 2116 \\ 
  Validation Accuracy & 81.3 & 78.5 & 78.5 \\ 
  Training Accuracy & 96.7 & 93.0 & 93.0 \\ 
    \hline
  $M-F1_{picture-BBCH}$ & 59.4 & 57.4 & 57.8 \\ 
  $M-F1_{parcel-BBCH}$ & 86.9 & 86.0 & 85.6 \\ 
  $M-F1_{picture-CROP}$ & 62.3 & 60.4 & 60.9 \\ 
  $M-F1_{parcel-CROP}$ & 88.1 & 88.2 & 87.5 \\ 
   \hline
\end{tabular}

\end{table}

\subsection{Phenology-level recognition}
\label{sec:bbchlevel}
\begin{figure*}
        \centering
        \begin{subfigure}[b]{0.8\textwidth}
            \centering
            \includegraphics[width=\textwidth]{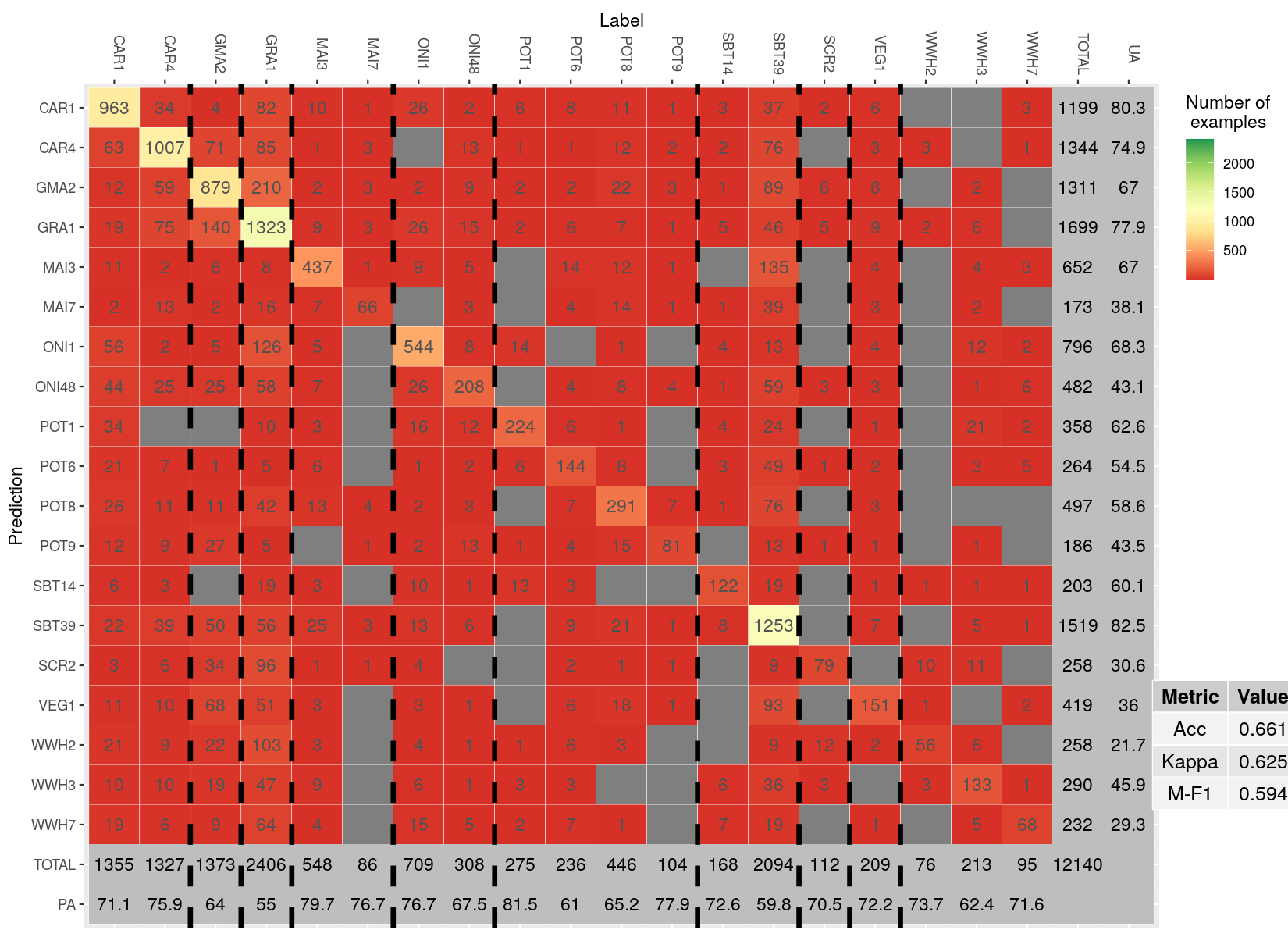}
            \caption[]%
            {{\small BBCH classes - picture level}}    
            \label{fig:cMatrixBBCH}
        \end{subfigure}\hspace{0.1\textwidth}%
        \begin{subfigure}[b]{0.8\textwidth}  
            \centering 
            \includegraphics[width=\textwidth]{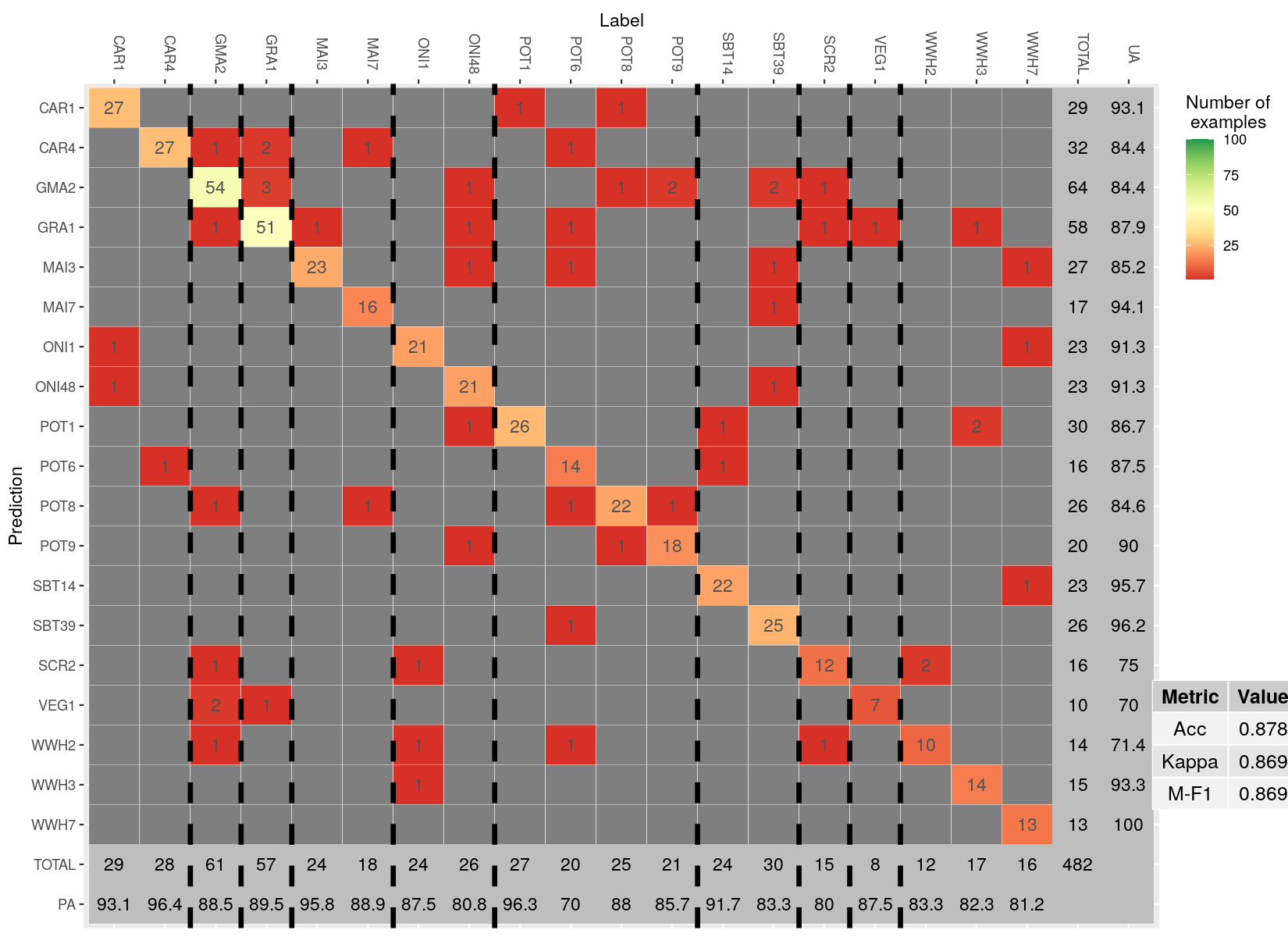}
            \caption[]%
            {{\small BBCH classes - parcel level}}    
            \label{fig:cMatrixBBCHParcels}
        \end{subfigure}
        \caption[]
        {\small Confusion matrices for phenology recognition (19 BBCH classes) at picture and parcel level. \ac{PA} and \ac{UA} are in the margins. See \ref{tab:allbbch_explained} for label details.} 
        \label{fig:bbchconfmatr}
    \end{figure*}

The best model (\#44) is applied on the independent inference set (12,140 pictures on 482 parcels). The confusion matrices are displayed in \ref{fig:bbchconfmatr} for both picture (\ref{fig:cMatrixBBCH}) and parcel (\ref{fig:cMatrixBBCHParcels}) level. 
As described in section \ref{sec:validation}, the bare soil ('BSO0') and other ('OTH0') classes are removed to assess the performance resulting in a 19-class confusion matrix. At the picture level, the model reached a \ac{M-F1} of \text{59.4\%}, while \text{86.9\%} was achieved when aggregating to parcel-level, an increase of \text{27.5\%}. At the image level, the model is certainly more likely to make more errors of commission than of omission, which is reflected by the relatively lower values of \ac{UA} compared to those of \ac{PA} (see \ref{fig:cMatrixBBCHParcels}. While \ac{PA} is, for the majority of classes, above 70 (excluding GMA2, GRA1, ONI48, POT6, POT8, SBT39, WWH3), the \ac{UA} score exceeds 70 for only 4 classes (CAR1, CAR4, GRA1, SBT39). Especially poor in this case are the cereal classes, spring cereal (SCR2) and \ac{WWH}, exhibiting confusion between themselves, and with \ac{GRA}. This is also reflected in the low \ac{UA} for these classes, meaning the model was miss-labelling many images from other classes as belonging to either of the cereal classes. Furthermore the distinction between cereals is difficult before the crop has elongated to form more recognisable canopy elements (e.g. ears). \ac{SCR} is actually a composite class used in the early surveys. Most parcels in these class evolve into distinct \ac{SBA} and \ac{SWH} in later surveys. Indeed it is the case that the class with the lowest \ac{PA}, GRA1, is the one that is most often confused with other classes. At the parcel level, however, the confusion is significantly less with \ac{PA} values in the high 80s and 90s. Interestingly, the \ac{UA} trend does not follow the same pattern at parcel level. For example, the class with the second to worst \ac{UA} at the picture level (WWH7 with \text{29.3}), has the best \ac{UA} on the parcel level (\text{100}).  

\subsection{Crop-level recognition}
\label{sec:clevel}

A total of 10 crops occur in the 19 crop-BBCH combinations. Performance at crop level is evaluated by running the model used to detect the phenological stages, and subsequently keeping the classification of the most likely crop class removing confusion in between the BBCH stages of the same crop. Model configuration 44 yielded a \ac{M-F1} of \text{62.3\%} on the picture level and \text{88.1\%} on the parcel level, meaning an increase of \text{25.8\%}. At picture level, there are both errors of commission and omission, reflected by the relatively low values of \ac{PA} and \ac{UA}. As with the parcel level performance on BBCH, the error decreases significantly from picture to parcel. The major part of the confusion seems to stem from the grass (GRA), green manure (GMA), and winter wheat (WWH) classes, whereby parcels from other crops get classified as either grass or winter wheat thus dropping the \ac{UA} of these classes; and green manure parcels get classified themselves as belonging to other classes, thereby lowering the \ac{PA} of this class.

\begin{figure*}
        \centering
        \begin{subfigure}[b]{0.475\textwidth}
            \centering
            \includegraphics[width=\textwidth]{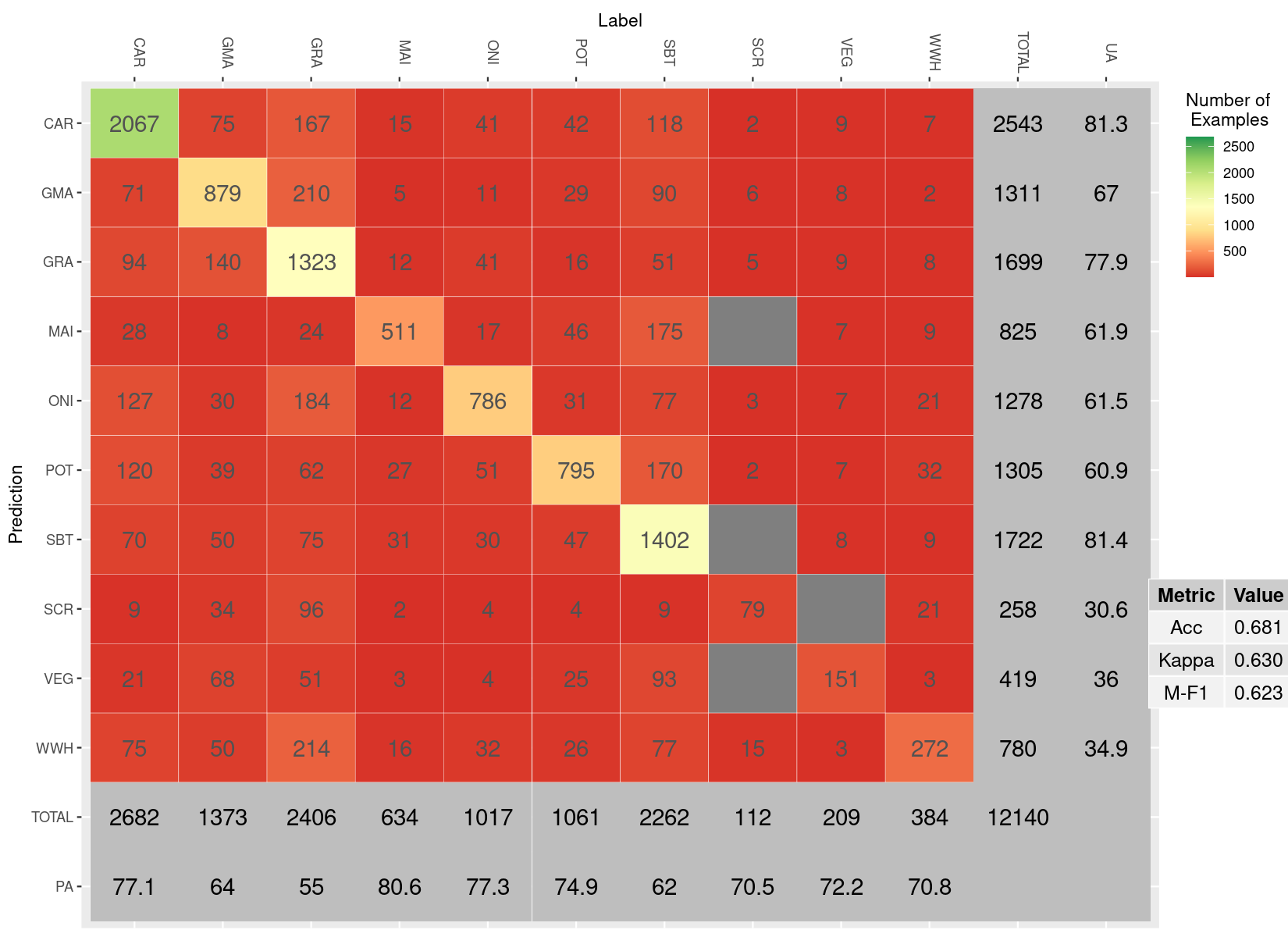}
            \caption[]%
            {{\small Crop classes - picture level}}    
            \label{fig:cMatrixCrop}
        \end{subfigure}
        \hfill
        \begin{subfigure}[b]{0.475\textwidth}  
            \centering 
            \includegraphics[width=\textwidth]{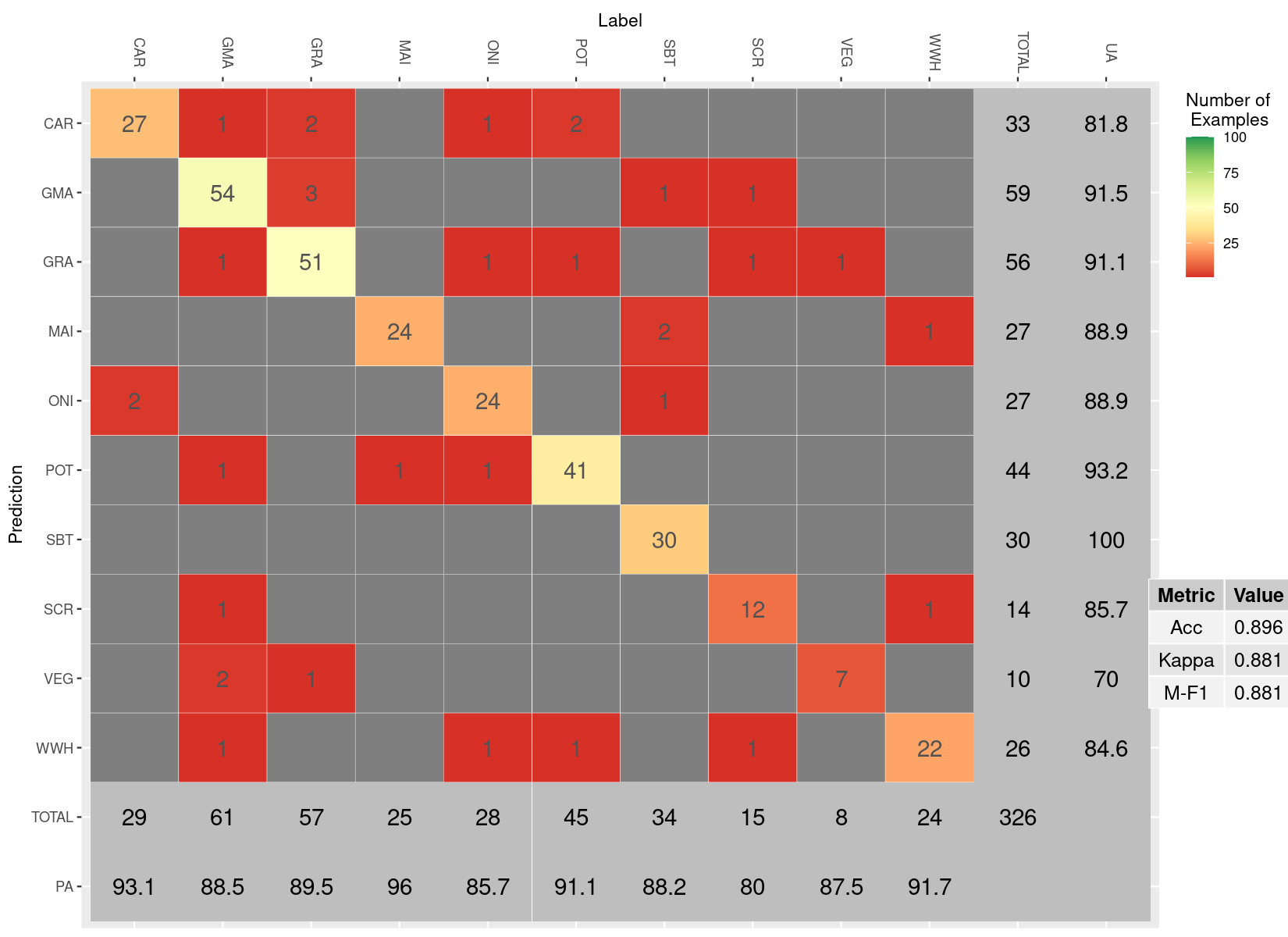}
            \caption[]%
            {{\small Crop classes - parcel level}}    
            \label{fig:cMatrixCropParcels}
        \end{subfigure}
        \caption[]
        {\small Confusion matrices for crop recognition (10 crop type) at (a) picture and (b) parcel level. \ac{UA} and \ac{PA} are in the margins.} 
        \label{fig:cropconfmatr}
    \end{figure*}

\section{Discussion}
\label{S:Discussion}

\subsection{Context}
Similar studies have recently been carried out although direct comparison is difficult as protocols and crops surveyed vary significantly. \citet{wu2021identification} obtained an overall accuracy of  91.1\% for 12 crop types (bare land, cotton, maize, peanut, rapeseed, rice, sorghum, soybean, sunflower, tobacco, vegetable, wheat) using majority voting for five classifier benchmarks. However, overall accuracy metrics are probably not the best for imbalanced datasets \citep{branco2015survey}. A study by \citet{yan2021exploring} explored Google Street View with deep learning for crop type mapping resulting in an accuracy ranging from 92\% for 8 crops (alfalfa, almonds, corn, cotton, grape, other, pisctachios, rice) to 97\% for 3 crops (corn, other, soybean). While the crops and metrics used are very different, the measured performance of \citet{wu2021identification} and \citet{yan2021exploring} are higher. Further studies and open datasets for benchmarking different methods are needed to cross-compare the results more consistently.

In any case, the study is a push towards a new state of the art in combining existing approaches into a methodology that can be easily scaled and executed for a reasonable cost. While other work has previously made contributions to plant recognition \citep{ringland2019characterization} and phenology recognition \citep{yalcin2015phenology}, and devising smart massive image-capture for computer vision \citep{wu2021identification}, there has not been a push towards having phenology level information collected with such methods. In terms of technical novelty, the work shows an automated method for a much easier and faster massive image labeling (section \ref{sec:in-situ_sample}). Additionally, the process for performing parcel-level classification through majority voting, multiple modes, and maximum probability summing, albeit simplistic, provides near-real-time parcel relevant information. This could be of major use to single or organizations of farmers, to agri-orientated consultancies, or to regulatory bodies.

\subsection{Limitations}
\label{S:limitations}

Although novel in theoretical, semantic, and technical regards, the study nevertheless suffers from some notable shortcomings and drawbacks. 

Firstly, in the proposed approach, the input data looses the majority of its detail due to the network's inability to handle full resolution images. It is well known that current deep learning architectures use input images that are of low resolution. For instance in radiography, the optimal input image size is between 256 and 448 pixels per dimension (\citet{sabottke2020effect}), and in general the trend is followed in the agricultural domain. The MobileNet V2 \citep{sandler2018mobilenetv2} used here automatically re-scales all input images to the largest resolution it can handle, which is 224 pixels per dimension. The full resolution images collected for this study are of dimensions 3840 pixels in width to 2160 pixels in height. Thus each image gets re-scaled from its parallelogram shape 17.1 times in width and 9.6 times in height to produce the rectangular input image that the network actually trains on. With such a huge reduction and loss of information, it is not surprising that much confusion exists between the relevant classes (see Section \ref{sec:recomend} for recommendations for future studies). Furthermore, due to the fact that the images are in full resolution, there is a lot of additional noise on the image, in the form of the sky, the road, grass or dirt margins around the fields, and additional objects in the distance. This can obfuscate the classification that is in principle solely based on the crop. 

In addition, further work could focus on bench-marking different families of networks and architectures, besides the flavours of MobileNet v2 considered here. This could include nets that use similar input resolutions (224x224) such as SqueezeNet (\citet{iandola2016squeezenet}) or ShuffleNet (\citet{zhang2018shufflenet}) or nets that take advantage of the high resolution SLI such as NASNet - NASNetLarge (\citet{zoph2018learning}) (331x331), the GPipe assisted training of AmoebaNet (480x480) (\citet{huang2019gpipe}), and finally the B7 version of EfficientNet (600x600) (\citet{tan2019efficientnet}). The advantage of these latter nets it that they can discriminate more features while the drawback is that they are not on-device and mobile-friendly.

Secondly, the way the workflow deals with the 'other' and 'bare soil' classes can be improved (Section \ref{sec:validation}). A total of 3,065 images were excluded consisting of 2,361 labeled 'bare soil' and 704 labeled 'other' respectively to facilitate accuracy calculations across models.
Excluding 'bare soil' images increased the performance of the vast majority of model configurations. For each performance level ($M-F1_{picture-BBCH}$, $M-F1_{parcelBBCH}$, $M-F1_{picture-crop}$, and $M-F1_{parcel-crop}$), the increase was respectively 1.9\%, 3.6\%, 2.3\%, and 5.5\%. This suggests that 'bare soil' is either significantly more present in the dataset and is ergo a disproportionately imbalanced class, or that it is too hard to be distinguished from seedlings. Nonetheless, models that correctly distinguished 'bare soil' from other classes were penalized. Further developments could consider hierarchically classifying bare soil and crop in a first step, and secondly distinguishing different types of bare soil e.g. related to ploughing patters associated to crops.

Thirdly, the study assumes a uniform distribution of a BBCH stage across the entire roadside perimeter of the parcel. In reality, the detailed BBCH observation was made by the surveyor at a particular point in the parcel. Due to the nature of the logic described in section \ref{sec:in-situ_sample}, the processing chain assumes that all images associated with the parcel exhibit the same stage. In doing so, we may be introducing some noise by grouping together pictures that actually show the crop in a variety of stages.

\subsection{Parcel-level interpretation}
Estimating parcel-level model performance is a key aspect of this study. Specifically, the questions are what is the minimum \ac{PPP} needed for a correct classification, and after how \ac{PPP} is a correct classification certain? 

Although the number of surveyed parcels is 307 (section \ref{sec:study_area}), the parcels over which results are calculated on BBCH and crop levels are 482 (\ref{fig:cMatrixBBCHParcels}) and 326 (\ref{fig:cMatrixCropParcels}) respectively. Rather than aggregating on the unique combination of crop type and survey date, the study chose to perform the aggregation based on crop type and parcel ID. The latter combination is not unique and thus parcel IDs get re-coded to reflect the different BBCH stages (for single crop parcels) and crop types (for multi-crop parcels) of the parcel. This method is preferable, as it deals simultaneously with cases of more prevalent BBCH stages (SBT39) and sparsely represented classes.

\begin{figure*}
\centering
\hspace*{-1cm}   
\includegraphics[width=0.8\textwidth]{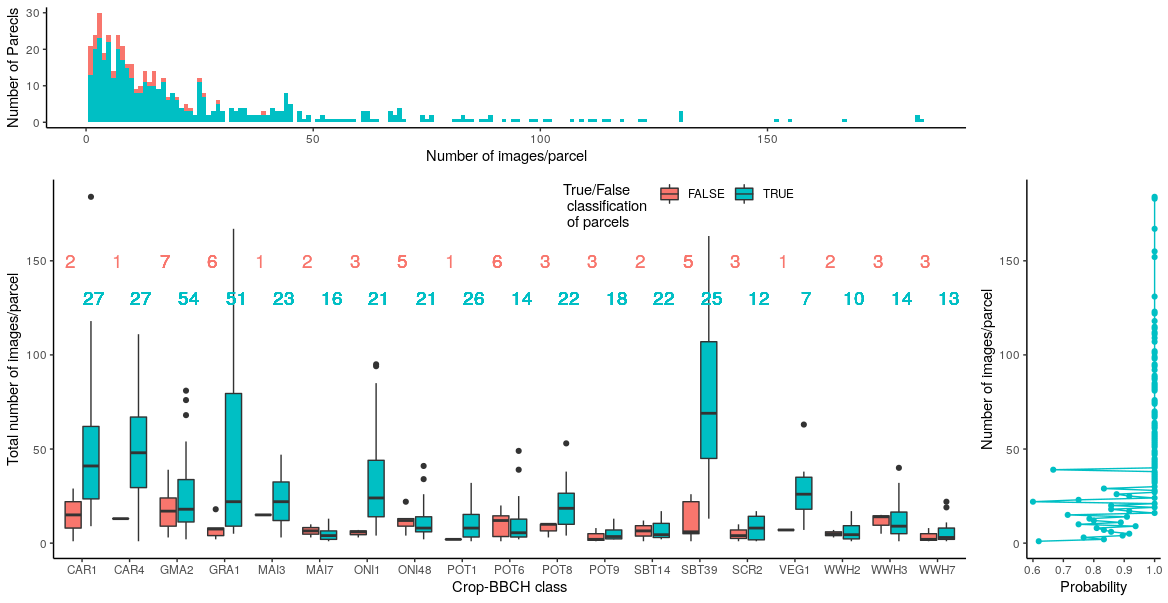}
\caption{Distribution of correctly and incorrectly classified parcels. The three plots are derived from the same table and show a different aspects of the data. The top plot is a histogram of the number of pictures per parcel, the scatter plot to the right shows the probability of a parcel with a certain number of pictures to be classified correctly, and the barplot in the middle shows a breakdown per class of the data with the numbers over each bar indicating the number of parcels that belong to each group of correctly or incorrectly classified parcels.}
\label{fig:barplotBBCH_hist_combined}
\end{figure*}

The histogram on top of \ref{fig:barplotBBCH_hist_combined} shows the frequency of correctly and incorrectly classified parcels in terms of the number of \ac{PPP}. The data is skewed to the right, meaning that the majority of incorrect and correct classifications occur at the lower ranges. The mean of incorrectly and correctly classified parcels is \text{9.6} and \text{27.4} \ac{PPP} respectively. It is thus impossible to answer the first question, as there are both correctly and incorrectly classified parcels with a single picture. Alternatively, the data gives insight into the probability for a correct classification of a parcel with a given \ac{PPP}. This information is shown on the connected scatter plot on the right of \ref{fig:barplotBBCH_hist_combined}. 
The histogram also illustrates the evolution of certainty in getting a correct classification with regards to natural breaks. In the range of 0-25 \ac{PPP}, the proportion of correct to incorrect parcel classifications are \text{83.4\%} and \text{16.6\%} respectively; in the range of 25-50 \ac{PPP} the percentages are \text{96\%} and \text{4\%}; finally for parcels with over 50 pictures the classifications are 100\% correct. 
The figure in the middle gives a more detailed breakdown of the other two. Firstly for almost all the classes, the range of values for correctly classified parcels is larger than for the incorrect ones, excluding POT6. However, for six out of nineteen classes (MAI7, ONI48, POT6, SBT14, WWH2, WWH3) the mean number of \ac{PPP} is higher for the incorrectly classified parcels. With such a spread between the anomalous classes it is impossible to attribute this to a specific crop or set thereof. From visually inspecting the locations of the miss-classified parcels, there is no spatial clustering. This suggests that, especially in the lower ranges of \ac{PPP} (below 25), image abundance cannot solely guarantee correct classifications, but rather that there is also a defining role for image quality.

\begin{figure*}
        \centering
        \begin{subfigure}[b]{0.475\textwidth}
            \centering
            \includegraphics[width=\textwidth]{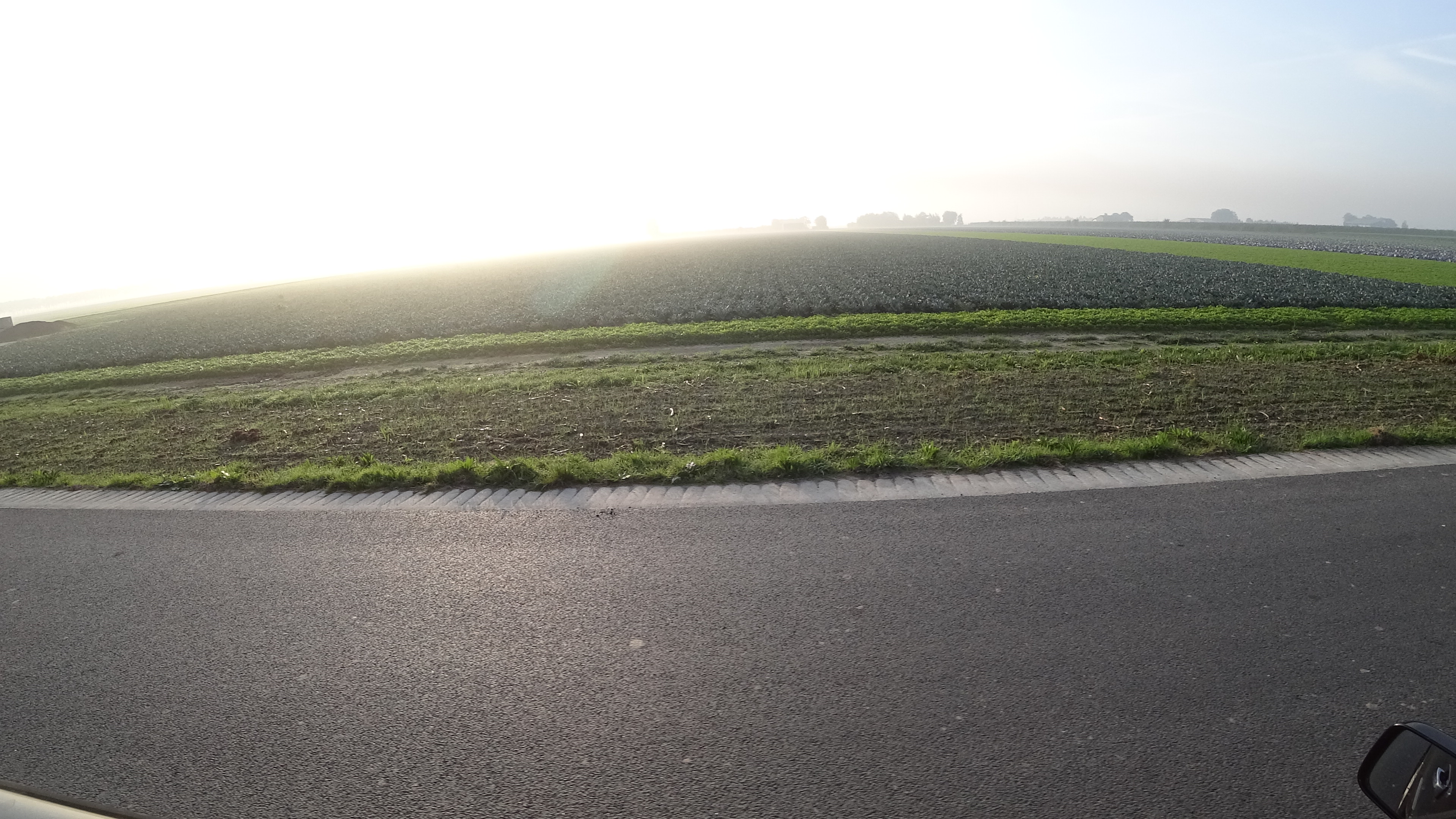}
            \caption[]%
            {{\small Sun glare}}    
            \label{fig:sunglare}
        \end{subfigure}
        \hfill
        \begin{subfigure}[b]{0.475\textwidth}  
            \centering 
            \includegraphics[width=\textwidth]{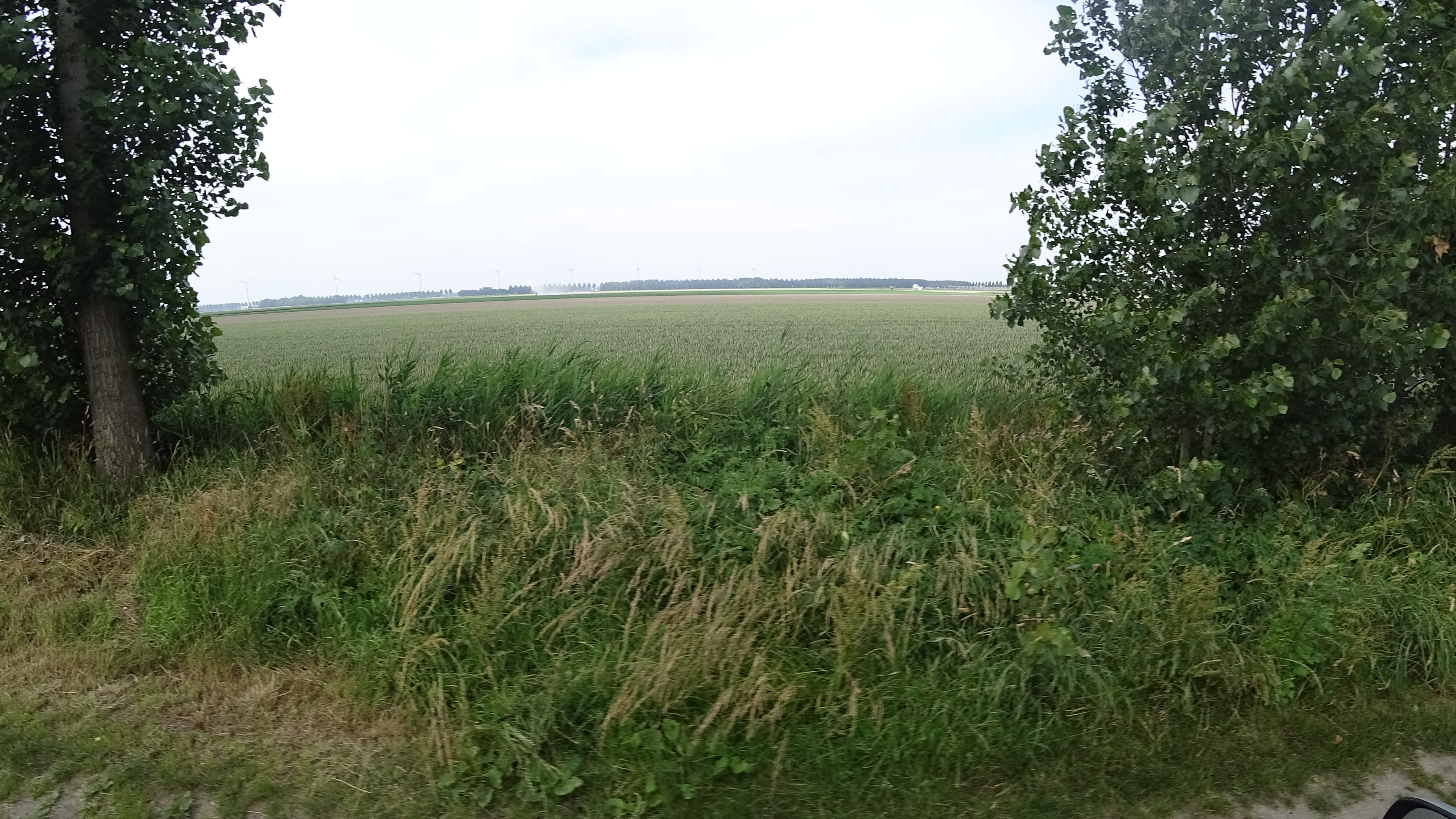}
            \caption[]%
            {{\small Tall grass}}    
            \label{fig:tallgrass}
        \end{subfigure}
        \vskip\baselineskip
        \begin{subfigure}[b]{0.475\textwidth}   
            \centering 
            \includegraphics[width=\textwidth]{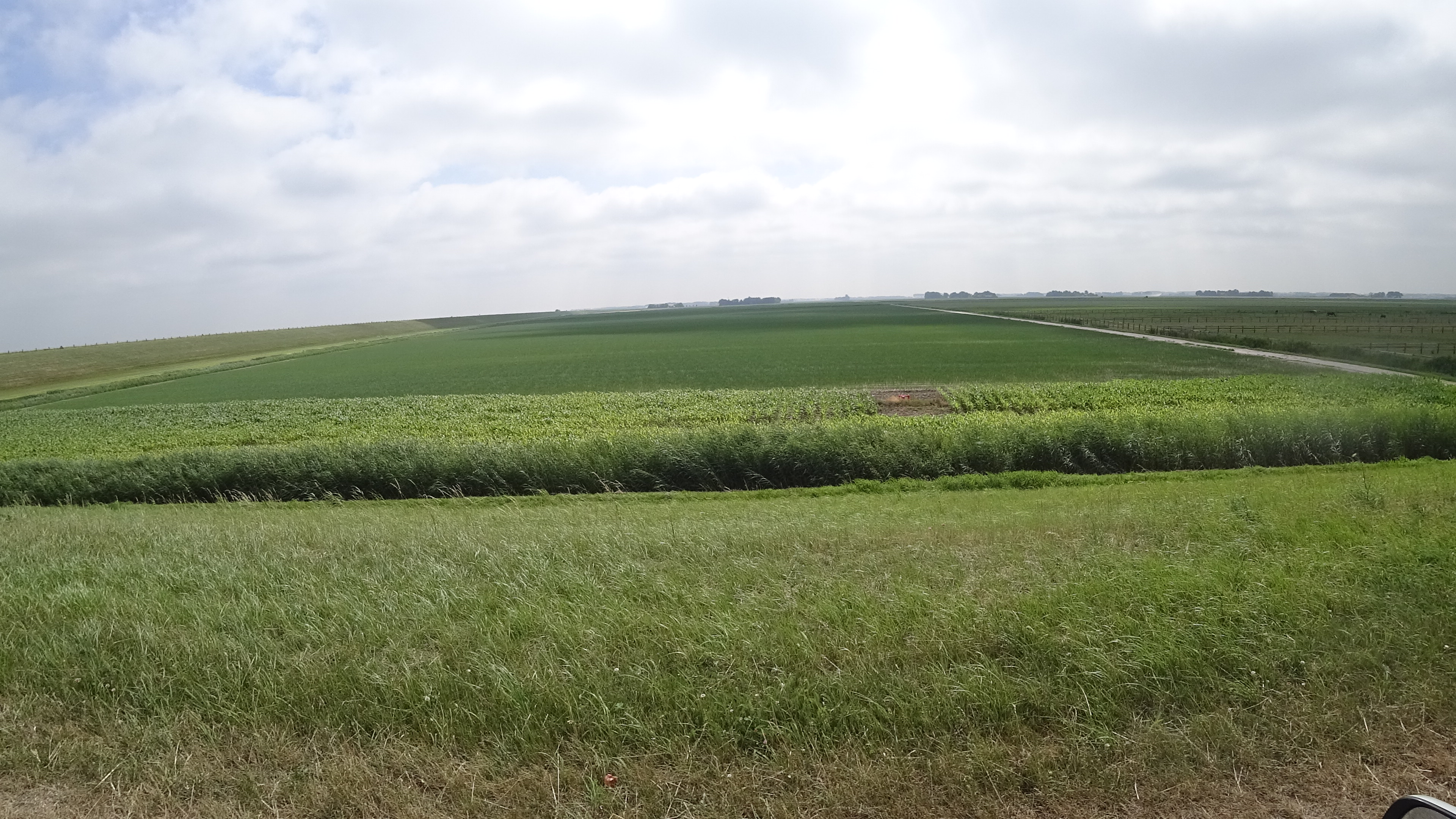}
            \caption[]%
            {{\small Parcel in distance}}    
            \label{fig:carfaraway}
        \end{subfigure}
        \hfill
        \begin{subfigure}[b]{0.475\textwidth}   
            \centering 
            \includegraphics[width=\textwidth]{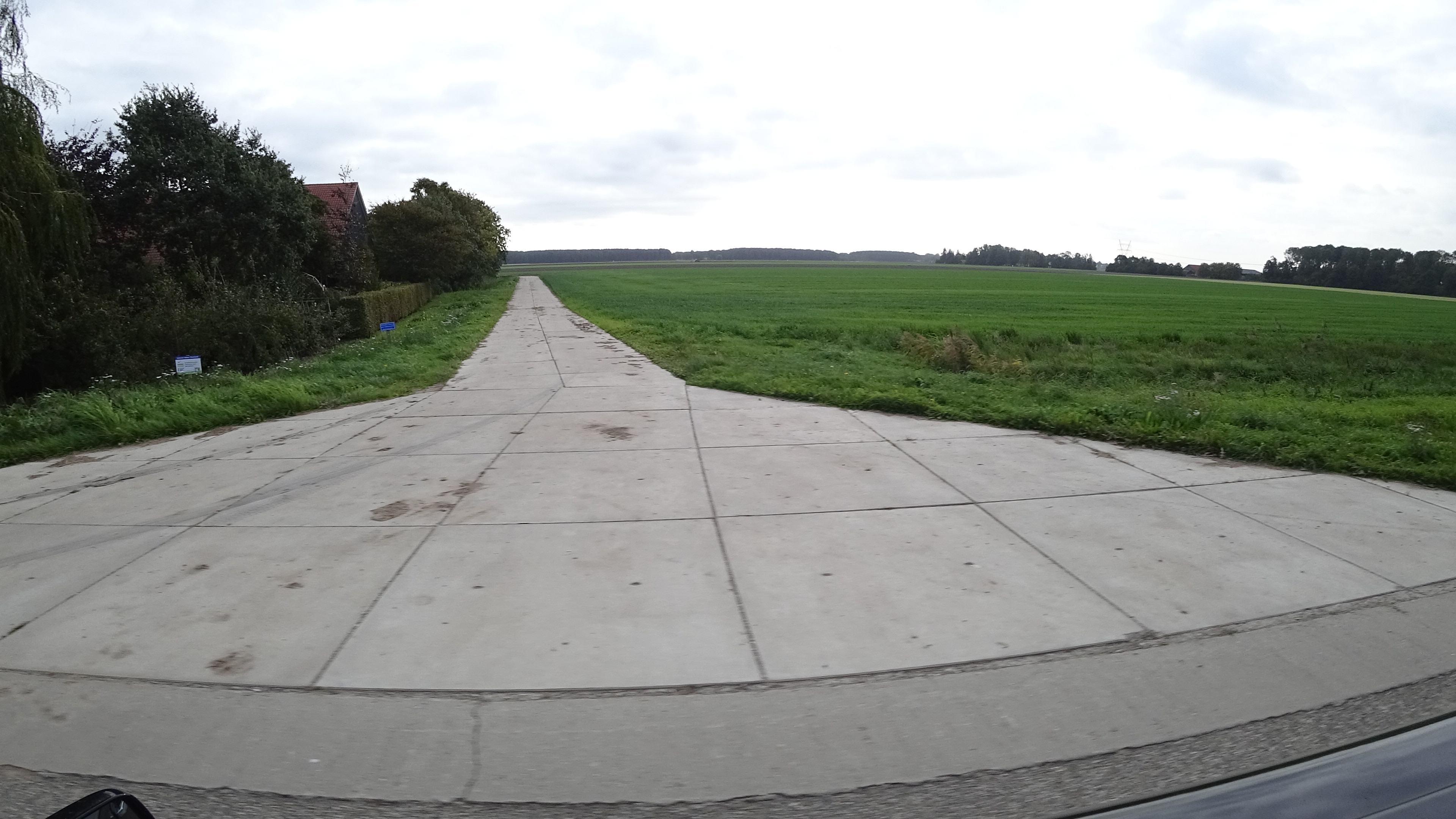}
            \caption[]%
            {{\small Faulty distance to centroid filter}}    
            \label{fig:baddtc}
        \end{subfigure}
        \vskip\baselineskip
                \begin{subfigure}[b]{0.475\textwidth}   
            \centering 
            \includegraphics[width=\textwidth]{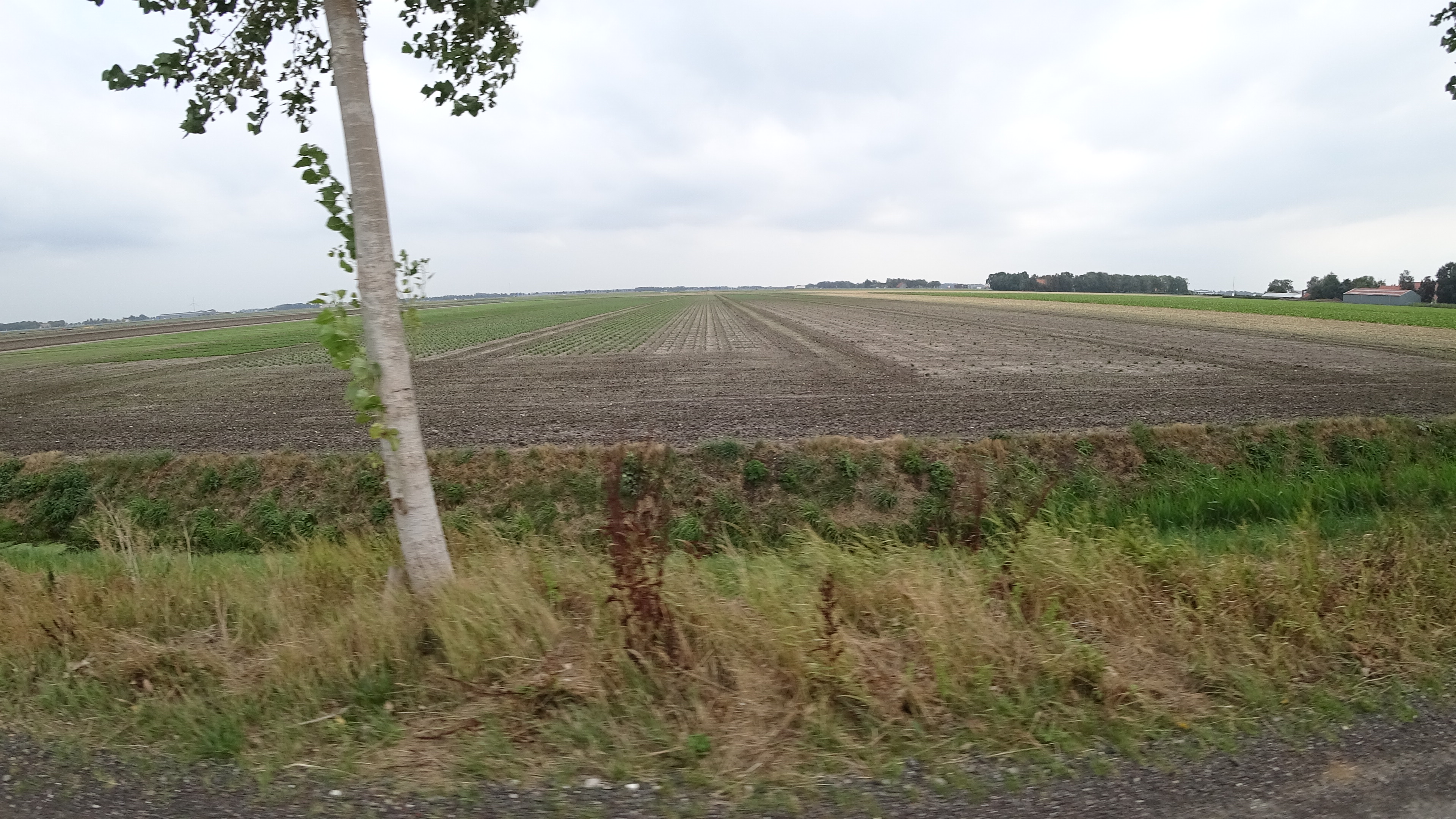}
            \caption[]%
            {{\small Potential label miss-match}}    
            \label{fig:dontlookright}
        \end{subfigure}
        \hfill
        \begin{subfigure}[b]{0.475\textwidth}   
            \centering 
            \includegraphics[width=\textwidth]{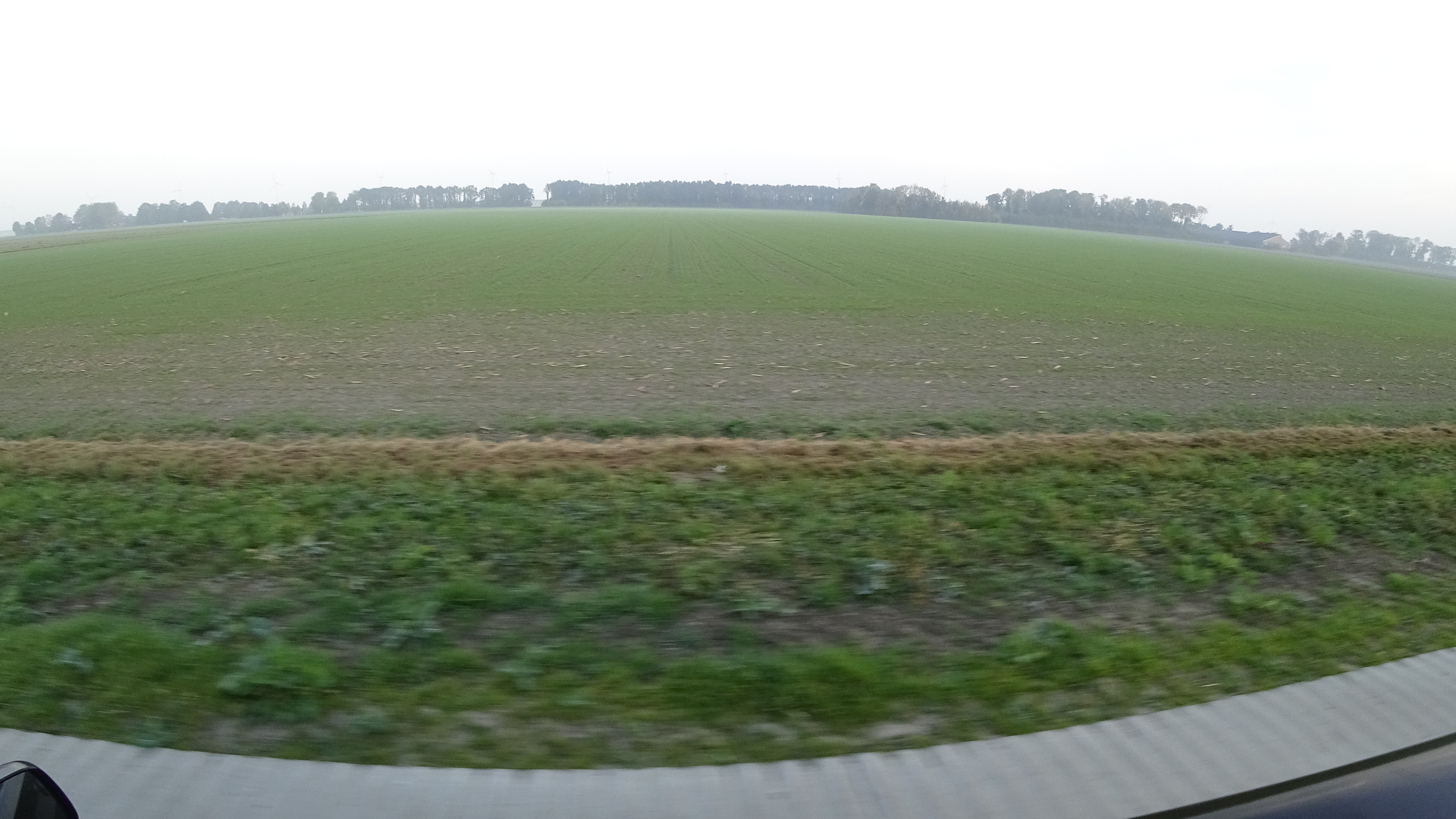}
            \caption[]%
            {{\small Blurred image}}    
            \label{fig:blurred}
        \end{subfigure}
        \caption[]
        {\small Illustrations from various miss-classified parcels of model 44 showcasing the different types of shortcomings that can exists, which make for an incorrect parcel classification.} 
    \end{figure*}

To gain insight into the role of image quality, the pictures of all 59 miss-classified parcels at BBCH level were visually inspected. The errors are due to environmental, geographic, or processing related factors. The first category includes issues like sun glare (\ref{fig:sunglare}), and tall grass or linear elements obstructing the view on the parcel (\ref{fig:tallgrass}). The second category includes situations where the camera is away from the parcel (\ref{fig:carfaraway}), and where parcels are imaged from different aspects. In the third category we find issues related to the distance to centroid ratio filter not performing adequately (\ref{fig:baddtc}), resulting in many pictures of parcel boundaries being included. In this class errors related to labelling are also included, an example is shown in \ref{fig:dontlookright} where parcel \text{2295159} is labelled as SBT39, but classified as ONI48. Comparing this example with the SBT39, as shown in \ref{fig:allcrops}, a clear difference is visible. Finally, we include here pictures that are blurred or out of focus (\ref{fig:blurred}). For parcels that are incorrectly classified, more than one factor could have led to an incorrect classification. A parcel may be both obscured by tall grass and also far away from the camera. Another point which comes up often is the relatively few pictures per parcel that are available. Although most of the miss-classified parcels exhibit one or more of these drawbacks, it must be noted that for a significant number of them (20 out of 59) there is nothing visibly wrong with the pictures and hence the incorrect classification is purely due to sub-optimal model performance.  

\subsection{Recommendations}
\label{sec:recomend}
The availability of frequent and high resolution Copernicus Sentinel satellite data streams requires complementary in-situ data for instantaneous cross-referencing and validation. To boost collection efficiency, novel ways to collect or derive such high quality reference data need to be explored. 
In-situ data collection is a tedious and expensive task and is often limited to small scales, and specific purposes, serving various scientific validation experiments. Only very limited amounts of standardized reference data end up in the open data domain, notwithstanding collaborative community efforts like the \ac{JECAM} \citep{kussul2014use}) of the \ac{GEO}. Efforts to use computer vision approaches on the archive of large scale standardized in-situ statistical surveys including geo-tagged pictures could provide new information on historical dynamics of land cover and land use. A recent harmonisation of the LUCAS \citep{d2020harmonised} survey has delivered 5.4 millions of land-cover-labeled pictures over 15 years statistically distributed in EU. More research is needed to apply state-of-the art computer vision methods on such datasets.

For the current study, tiling the full resolution images into 224x224 pure land cover tiles should significantly improve the model's performance. The challenge is to detect the parcel of interest in the landscape \ac{SLI}. This could be done by detecting the horizon line and then taking a buffer in a down-ward direction from the horizon to a relevant threshold that would encompass the pixels exhibiting only the parcel. This buffered group of pixels would then be split into 224x224 tiles in order to create better input images for the network. Alternatively, specific semantic segmentation nets such as Detectron2  \citep{wu2019detectron2} could be trained to detect the parcel and tiles could be extracted in a similar manner from the parcel segmentation mask. Such tiling would remove the problem of mixed classes inherent to the \ac{SLI} and would increase the number of pictures available as many tiles could be extracted from one \ac{SLI}.

Last but not least, recent  methodological developments for botanical taxonomic inventory using massive amount of pictures collected by crowdsourcing and experts such as PlantNet
\citep{affouard2017pl} could provide method, tools (i.e. API) taking benefit of a wide community. These methods and tools should be investigated for classification of \ac{SLI}.

\section{Conclusions}
In this paper, we have presented a framework to collect and extract crop type and phenological information from street level imagery using computer vision. We have shown how, during the 2018 growing season, high definition pictures were captured with side-looking action cameras in the Flevoland province of the Netherlands. Thanks to a classification performed with transfer learning of  convolutional neural network modules along with a an hypertuning methodology, we discriminate crop type with a Macro F1 score of 88.1\% and main phenological stage at 86.9\% at the parcel level. While a lot of developments are still needed to improve the performance of the classification, the proposed framework is a first step in speeding up high quality in-situ data collection and opens avenues for massive data collection via automated classification using computer vision.

\section{Author contributions}
R.D, M.Y., and M. V. conceptualized the study and designed the methodology, R.D., M.Y., L. M. processed the data. R.D., M.Y., L.M.,  M. V.  analyzed the data and wrote the paper.

\section{Acknowledgements}
The authors thank Guido Lemoine for his support and agronomic inputs. The authors would like to thank TerraSphere Imaging for the high quality data collection, in particular Menno de Vries and  Paul van der Voet. 
The authors would also like to thank Dominique Fasbender for his precious statistical advices and the JRC colleagues from the Big Data Analytics project for their support. The author also thank the Exploratory Research program of the Joint Research Centre for their support within the Rural Refocus project (31280).

\section{References}

\bibliography{sample.bib}

\appendix

\onecolumn

\clearpage

\tableofcontents

\setlength{\cftfignumwidth}{1.4cm}
\setlength{\cfttabnumwidth}{1.5cm}

\listoffigures
\listoftables

\clearpage

\setcounter{figure}{0}
\setcounter{table}{0}
\setcounter{page}{1}

\section*{Supplementary Material}

\label{AppendixA}
\captionsetup{list=no}
\renewcommand{\thetable}{Supplementary Table S\arabic{table}}
\renewcommand{\thefigure}{Supplementary Fig. S\arabic{figure}}


\begin{table}[ht]
\caption{Action camera settings for \ac{SLI} collection.}
\label{tab:sonycamsett}
\centering
\footnotesize
\begin{tabular}{ll}
  \hline
Settings & Used parameters \\ 
  \hline
Zoom & ON – 1.1 \\ 
Shooting & Time-lapse \\ 
Time lapse shooting interval & 1 sec \\ 
 Loop recording mode & Continue \\ 
Shooting scene settings & SCN N \\ 
 Angle & Wide \\ 
White balance & Automatic \\ 
Color mode & Natural \\ 
 Automatic exposure & AE – T \\ 
   \hline
\end{tabular}
\end{table}

\begin{footnotesize}
\begin{longtable}{ p{.21\textwidth} |  p{.15\textwidth} |  p{.10\textwidth} |  p{.40\textwidth} } 

\caption{Table explaining the meaning behind all the present BBCH stages of all surveyed crops. The table can be read as a detailed view of \ref{fig:bbchconfmatr}.} 
\label{tab:allbbch_explained}
\footnotesize
\\
\hline
\textbf{Crop BBCH } & \textbf{Crop common name} & \textbf{BBCH} & \textbf{Description} \\ 
\hline
\endfirsthead

\multicolumn{4}{c}%
{\tablename\ \thetable\ -- \textit{Continued from previous page}} \\
\hline
\textbf{Crop BBCH } & \textbf{Crop common name} & \textbf{BBCH} & \textbf{Description} \\ 
\hline
\endhead

\hline \multicolumn{2}{c}{\textit{Continued on next page}} \\
\endfoot

\hline
\endlastfoot

BSO0 & Bare soil &   0 & Untreated bar \\ 
  BSO1 & Bare soil &   1 & Untreated stubble \\ 
  BSO2 & Bare soil &   2 & Green manure \\ 
  BSO3 & Bare soil &   3 & Ploughed/converted \\ 
  BSO4 & Bare soil &   4 & Harrowed \\ 
  BSO5 & Bare soil &   5 & Ploughed and harrowed \\ 
  BSO6 & Bare soil &   6 & Seedbed prepared code \\ 
  \hline
  CAR0 & Carrot &   0 & Germination \\ 
  CAR1 & Carrot &   1 & Leaf development \\ 
  CAR2 & Carrot &   2 & No info \\ 
  CAR3 & Carrot &   3 & No info \\ 
  CAR4 & Carrot &   4 & Development of harvestable vegetative plant parts \\ 
  \hline
  GMA0 & Green manure &   0 & No info \\ 
  GMA1 & Green manure &   1 & No info \\ 
  GMA2 & Green manure &   2 & No info \\ 
  GMA3 & Green manure &   3 & No info \\ 
  GMA4 & Green manure &   4 & No info \\ 
  \hline
  GRA0 & Grassland &   0 & Bare soil \\ 
  GRA1 & Grassland &   1 & After mowing height 0-5cm \\ 
  GRA2 & Grassland &   2 & Leaves not yet bending, height 5-20cm \\ 
  GRA3 & Grassland &   3 & Before mowing, bending (height 20-30cm) \\ 
  \hline
  GRS7 & Grass seeds &   7 & No info \\ 
  \hline
  MAI0 & Maize &   0 & Germination \\ 
  MAI1 & Maize &   1 & Leaf development \\ 
  MAI3 & Maize &   3 & Stem elongation \\ 
  MAI5 & Maize &   5 & Inflorescence emergence, heading \\ 
  MAI6 & Maize &   6 & Flowering, anthesis \\ 
  MAI7 & Maize &   7 & Development of fruit \\ 
  MAI8 & Maize &   8 & Ripening code \\ 
  \hline
  ONI0 & Onion &   0 & Germination \\ 
  ONI1 & Onion &   1 & Leaf development (main shoot) \\ 
  ONI41 & Onion &  41 & Leaf bases begin to thicken or extend \\ 
  ONI45 & Onion &  45 & 50\% of the expected bulb or shaft diameter reached \\ 
  ONI48 & Onion &  48 & Leaves bent over in 50\% of plants \\ 
  ONI49 & Onion &  49 & Leaves dead, bulb top dry; dormancy \\ 
  \hline
  POT0 & Potato &   0 & Germination \\ 
  POT1 & Potato &   1 & Leaf development \\ 
  POT2 & Potato &   2 & Formation of basal side shoots below and above soil surface (main stem) \\ 
  POT3 & Potato &   3 & Main stem elongation (crop cover) \\ 
  POT5 & Potato &   5 & Inflorescence (cyme) emergence \\ 
  POT6 & Potato &   6 & Flowering \\ 
  POT8 & Potato &   8 & Ripening of fruit and seed \\ 
  POT9 & Potato &   9 & Senescence code \\ 
  \hline
  WCR0 & Cereal &   0 & Germination \\ 
  WCR1, SCR1 & Cereal &   1 & Leaf development \\ 
  SCR2, WWH2 & Cereal &   2 & Tillering \\ 
  SBA3, SCR3, WWH3 & Cereal &   3 & Stem elongation \\ 
  SBA5, SWH5, WWH5 & Cereal &   5 & Inflorescence emergence, heading \\ 
  SBA7, SWH7, WWH7 & Cereal &   7 & Development of fruit \\ 
  SBA8, SWH8, WWH8, WBA8 & Cereal &   8 & Ripening \\ 
  SBA9, SWH9, WWH9 & Cereal &   9 & Senescence code \\ 
  \hline
  SBT0 & Sugar beet &   0 & Germination \\ 
  SBT11 & Sugar beet &  11 & Leaf development -first leaf unfolded \\ 
  SBT14 & Sugar beet &  14 & Leaf development - 3 leaves unfolded \\ 
  SBT19 & Sugar beet &  19 & Leaf development - 9 or more leaves unfolded \\ 
  SBT2 & Sugar beet &   2 & No info \\ 
  SBT32 & Sugar beet &  32 & Leaves cover 20\% of ground \\ 
  SBT39 & Sugar beet &  39 & Crop cover complete: leaves cover 90\% of ground code \\ 
  \hline
  TLP0 & Tulip &   0 & No info \\ 
  TLP12 & Tulip &  12 & No info \\ 
  TLP45 & Tulip &  45 & No info \\ 
  TLP48 & Tulip &  48 & No info \\ 
  TLP49 & Tulip &  49 & No info \\ 
  TLP6 & Tulip &   6 & No info \\ 
  \hline
  VEG1 & Vegatable &   1 & Leaf development (Main shoot) \\ 
  VEG2 & Vegatable &   2 & Formation of side shoots \\ 
  VEG8 & Vegatable &   8 & Ripening of fruit and seed \\ 
  \hline

\end{longtable}
\end{footnotesize}

\end{document}

%% file: acrolist.tex
\acrodef{AUC}{Area Under the Curve }
\acrodef{AWS}{Amazon Web Services}

\acrodef{BDAP}{“Big Data Analytics Platform”}
\acrodef{BRP}{'Basisregistratie Gewaspercelen'}
\acrodef{BSO}{Bare Soil}

\acrodef{CNN}{Convolutional Neural Networks}

\acrodef{F1}{F1 Score}
\acrodef{FPR}{False Positive Rate}

\acrodef{GEO}{Group on Earth Observation}

\acrodef{JEODPP}{Joint Research Center Earth Observation Data and Processing Platform}
\acrodef{JECAM}{Joint Experiment of Crop Assessment and Monitoring}
\acrodef{JRC}{Joint Research Centre}

\acrodef{M-F1}{Macro F1 Score}

\acrodef{PA}{Producer Accuracy}

\acrodef{UA}{User Accuracy}

\acrodef{ROC}{Receiver-Operating Curve}

\acrodef{SLI}{Street Level Imagery}

\acrodef{TPR}{True Positive Rate}

\acrodef{PPP}{Pictures Per Parcel}

\acrodef{CAR}{Carrot}
\acrodef{GMA}{Green Manure}
\acrodef{GRA}{Grassland}
\acrodef{GRS}{Grass Seeds}
\acrodef{MAI}{Maize}
\acrodef{ONI}{Onion}
\acrodef{POT}{Potato}
\acrodef{SBA}{Summer Barley}
\acrodef{SBT}{Sugar Beet}
\acrodef{SCR}{Spring Cereals}
\acrodef{SWH}{Spring Wheat}
\acrodef{TLP}{Tulip}
\acrodef{VEG}{Vegetable}
\acrodef{WBA}{Winter Barley}
\acrodef{WCR}{Winter Cereals}
\acrodef{WWH}{Winter Wheat}
\acrodef{OTH}{Other}

%% file: FlevoVision-v1.bbl
\begin{thebibliography}{57}
\expandafter\ifx\csname natexlab\endcsname\relax\def\natexlab#1{#1}\fi
\providecommand{\url}[1]{\texttt{#1}}
\providecommand{\href}[2]{#2}
\providecommand{\path}[1]{#1}
\providecommand{\DOIprefix}{doi:}
\providecommand{\ArXivprefix}{arXiv:}
\providecommand{\URLprefix}{URL: }
\providecommand{\Pubmedprefix}{pmid:}
\providecommand{\doi}[1]{\href{http://dx.doi.org/#1}{\path{#1}}}
\providecommand{\Pubmed}[1]{\href{pmid:#1}{\path{#1}}}
\providecommand{\bibinfo}[2]{#2}
\ifx\xfnm\relax \def\xfnm[#1]{\unskip,\space#1}\fi
\bibitem[{Affouard et~al.(2017)Affouard, Go{\"e}au, Bonnet, Lombardo \&
  Joly}]{affouard2017pl}
\bibinfo{author}{Affouard, A.}, \bibinfo{author}{Go{\"e}au, H.},
  \bibinfo{author}{Bonnet, P.}, \bibinfo{author}{Lombardo, J.-C.}, \&
  \bibinfo{author}{Joly, A.} (\bibinfo{year}{2017}).
\newblock \bibinfo{title}{Plantnet app in the era of deep learning}.
\newblock In {\it \bibinfo{booktitle}{ICLR: International Conference on
  Learning Representations}\/}.
\bibitem[{Anami et~al.(2020)Anami, Malvade \& Palaiah}]{anami2020deep}
\bibinfo{author}{Anami, B.~S.}, \bibinfo{author}{Malvade, N.~N.}, \&
  \bibinfo{author}{Palaiah, S.} (\bibinfo{year}{2020}).
\newblock \bibinfo{title}{Deep learning approach for recognition and
  classification of yield affecting paddy crop stresses using field images}.
\newblock {\it \bibinfo{journal}{Artificial Intelligence in Agriculture}\/},
  {\it \bibinfo{volume}{4}\/}, \bibinfo{pages}{12--20}.
\bibitem[{Barve(2014)}]{barve2014discovering}
\bibinfo{author}{Barve, V.} (\bibinfo{year}{2014}).
\newblock \bibinfo{title}{Discovering and developing primary biodiversity data
  from social networking sites: A novel approach}.
\newblock {\it \bibinfo{journal}{Ecological Informatics}\/},  {\it
  \bibinfo{volume}{24}\/}, \bibinfo{pages}{194--199}.
\bibitem[{Branco et~al.(2015)Branco, Torgo \& Ribeiro}]{branco2015survey}
\bibinfo{author}{Branco, P.}, \bibinfo{author}{Torgo, L.}, \&
  \bibinfo{author}{Ribeiro, R.} (\bibinfo{year}{2015}).
\newblock \bibinfo{title}{A survey of predictive modelling under imbalanced
  distributions}.
\newblock {\it \bibinfo{journal}{arXiv preprint arXiv:1505.01658}\/}, .
\bibitem[{Cao et~al.(2021)Cao, Sun, Jiang, Li \& Xin}]{cao2021identifying}
\bibinfo{author}{Cao, M.}, \bibinfo{author}{Sun, Y.}, \bibinfo{author}{Jiang,
  X.}, \bibinfo{author}{Li, Z.}, \& \bibinfo{author}{Xin, Q.}
  (\bibinfo{year}{2021}).
\newblock \bibinfo{title}{Identifying leaf phenology of deciduous broadleaf
  forests from phenocam images using a convolutional neural network regression
  method}.
\newblock {\it \bibinfo{journal}{Remote Sensing}\/},  {\it
  \bibinfo{volume}{13}\/}, \bibinfo{pages}{2331}.
\bibitem[{Champ et~al.(2020)Champ, Mora-Fallas, Go{\"e}au, Mata-Montero, Bonnet
  \& Joly}]{champ2020instance}
\bibinfo{author}{Champ, J.}, \bibinfo{author}{Mora-Fallas, A.},
  \bibinfo{author}{Go{\"e}au, H.}, \bibinfo{author}{Mata-Montero, E.},
  \bibinfo{author}{Bonnet, P.}, \& \bibinfo{author}{Joly, A.}
  (\bibinfo{year}{2020}).
\newblock \bibinfo{title}{Instance segmentation for the fine detection of crop
  and weed plants by precision agricultural robots}.
\newblock {\it \bibinfo{journal}{Applications in plant sciences}\/},  {\it
  \bibinfo{volume}{8}\/}, \bibinfo{pages}{e11373}.
\bibitem[{Cloud(2011)}]{cloud2011amazon}
\bibinfo{author}{Cloud, A. E.~C.} (\bibinfo{year}{2011}).
\newblock \bibinfo{title}{Amazon web services}.
\newblock {\it \bibinfo{journal}{Retrieved November}\/},  {\it
  \bibinfo{volume}{9}\/}, \bibinfo{pages}{2011}.
\bibitem[{d'Andrimont et~al.(2018)d'Andrimont, Lemoine \& Van~der
  Velde}]{d2018targeted}
\bibinfo{author}{d'Andrimont, R.}, \bibinfo{author}{Lemoine, G.}, \&
  \bibinfo{author}{Van~der Velde, M.} (\bibinfo{year}{2018}).
\newblock \bibinfo{title}{Targeted grassland monitoring at parcel level using
  sentinels, street-level images and field observations}.
\newblock {\it \bibinfo{journal}{Remote Sensing}\/},  {\it
  \bibinfo{volume}{10}\/}, \bibinfo{pages}{1300}.
\bibitem[{d'Andrimont et~al.(2020{\natexlab{a}})d'Andrimont, Taymans, Lemoine,
  Ceglar, Yordanov \& van~der Velde}]{d2020detecting}
\bibinfo{author}{d'Andrimont, R.}, \bibinfo{author}{Taymans, M.},
  \bibinfo{author}{Lemoine, G.}, \bibinfo{author}{Ceglar, A.},
  \bibinfo{author}{Yordanov, M.}, \& \bibinfo{author}{van~der Velde, M.}
  (\bibinfo{year}{2020}{\natexlab{a}}).
\newblock \bibinfo{title}{Detecting flowering phenology in oil seed rape
  parcels with sentinel-1 and-2 time series}.
\newblock {\it \bibinfo{journal}{Remote sensing of environment}\/},  {\it
  \bibinfo{volume}{239}\/}, \bibinfo{pages}{111660}.
\bibitem[{d'Andrimont et~al.(2020{\natexlab{b}})d'Andrimont, Yordanov,
  Martinez-Sanchez, Eiselt, Palmieri, Dominici, Gallego, Reuter, Joebges,
  Lemoine et~al.}]{d2020harmonised}
\bibinfo{author}{d'Andrimont, R.}, \bibinfo{author}{Yordanov, M.},
  \bibinfo{author}{Martinez-Sanchez, L.}, \bibinfo{author}{Eiselt, B.},
  \bibinfo{author}{Palmieri, A.}, \bibinfo{author}{Dominici, P.},
  \bibinfo{author}{Gallego, J.}, \bibinfo{author}{Reuter, H.~I.},
  \bibinfo{author}{Joebges, C.}, \bibinfo{author}{Lemoine, G.} et~al.
  (\bibinfo{year}{2020}{\natexlab{b}}).
\newblock \bibinfo{title}{Harmonised lucas in-situ land cover and use database
  for field surveys from 2006 to 2018 in the european union}.
\newblock {\it \bibinfo{journal}{Scientific Data}\/},  {\it
  \bibinfo{volume}{7}\/}, \bibinfo{pages}{1--15}.
\bibitem[{Deng et~al.(2009)Deng, Dong, Socher, Li, Li \&
  Fei-Fei}]{deng2009imagenet}
\bibinfo{author}{Deng, J.}, \bibinfo{author}{Dong, W.},
  \bibinfo{author}{Socher, R.}, \bibinfo{author}{Li, L.-J.},
  \bibinfo{author}{Li, K.}, \& \bibinfo{author}{Fei-Fei, L.}
  (\bibinfo{year}{2009}).
\newblock \bibinfo{title}{Imagenet: A large-scale hierarchical image database}.
\newblock In {\it \bibinfo{booktitle}{2009 IEEE conference on computer vision
  and pattern recognition}\/} (pp. \bibinfo{pages}{248--255}).
\newblock \bibinfo{organization}{Ieee}.
\bibitem[{Deus et~al.(2016)Deus, Silva, Catry, Rocha \&
  Moreira}]{deus2016google}
\bibinfo{author}{Deus, E.}, \bibinfo{author}{Silva, J.~S.},
  \bibinfo{author}{Catry, F.~X.}, \bibinfo{author}{Rocha, M.}, \&
  \bibinfo{author}{Moreira, F.} (\bibinfo{year}{2016}).
\newblock \bibinfo{title}{Google street view as an alternative method to car
  surveys in large-scale vegetation assessments}.
\newblock {\it \bibinfo{journal}{Environmental Monitoring and Assessment}\/},
  {\it \bibinfo{volume}{188}\/}, \bibinfo{pages}{1--14}.
\bibitem[{ElQadi et~al.(2017)ElQadi, Dorin, Dyer, Burd, Bukovac \&
  Shrestha}]{elqadi2017mapping}
\bibinfo{author}{ElQadi, M.~M.}, \bibinfo{author}{Dorin, A.},
  \bibinfo{author}{Dyer, A.}, \bibinfo{author}{Burd, M.},
  \bibinfo{author}{Bukovac, Z.}, \& \bibinfo{author}{Shrestha, M.}
  (\bibinfo{year}{2017}).
\newblock \bibinfo{title}{Mapping species distributions with social media
  geo-tagged images: case studies of bees and flowering plants in australia}.
\newblock {\it \bibinfo{journal}{Ecological informatics}\/},  {\it
  \bibinfo{volume}{39}\/}, \bibinfo{pages}{23--31}.
\bibitem[{Gebru et~al.(2017)Gebru, Krause, Wang, Chen, Deng, Aiden \&
  Fei-Fei}]{gebru2017using}
\bibinfo{author}{Gebru, T.}, \bibinfo{author}{Krause, J.},
  \bibinfo{author}{Wang, Y.}, \bibinfo{author}{Chen, D.},
  \bibinfo{author}{Deng, J.}, \bibinfo{author}{Aiden, E.~L.}, \&
  \bibinfo{author}{Fei-Fei, L.} (\bibinfo{year}{2017}).
\newblock \bibinfo{title}{Using deep learning and google street view to
  estimate the demographic makeup of neighborhoods across the united states}.
\newblock {\it \bibinfo{journal}{Proceedings of the National Academy of
  Sciences}\/},  {\it \bibinfo{volume}{114}\/}, \bibinfo{pages}{13108--13113}.
\bibitem[{Go{\"e}au et~al.(2018)Go{\"e}au, Joly, Bonnet, Lasseck, {\v{S}}ulc \&
  Hang}]{goeau2018deep}
\bibinfo{author}{Go{\"e}au, H.}, \bibinfo{author}{Joly, A.},
  \bibinfo{author}{Bonnet, P.}, \bibinfo{author}{Lasseck, M.},
  \bibinfo{author}{{\v{S}}ulc, M.}, \& \bibinfo{author}{Hang, S.~T.}
  (\bibinfo{year}{2018}).
\newblock \bibinfo{title}{Deep learning for plant identification: how the web
  can compete with human experts}.
\newblock {\it \bibinfo{journal}{Biodiversity Information Science and
  Standards}\/},  {\it \bibinfo{volume}{2}\/}, \bibinfo{pages}{e25637}.
\bibitem[{Howard et~al.(2017)Howard, Zhu, Chen, Kalenichenko, Wang, Weyand,
  Andreetto \& Adam}]{howard2017mobilenets}
\bibinfo{author}{Howard, A.~G.}, \bibinfo{author}{Zhu, M.},
  \bibinfo{author}{Chen, B.}, \bibinfo{author}{Kalenichenko, D.},
  \bibinfo{author}{Wang, W.}, \bibinfo{author}{Weyand, T.},
  \bibinfo{author}{Andreetto, M.}, \& \bibinfo{author}{Adam, H.}
  (\bibinfo{year}{2017}).
\newblock \bibinfo{title}{Mobilenets: Efficient convolutional neural networks
  for mobile vision applications}.
\newblock {\it \bibinfo{journal}{arXiv preprint arXiv:1704.04861}\/}, .
\bibitem[{Huang et~al.(2019)Huang, Cheng, Bapna, Firat, Chen, Chen, Lee, Ngiam,
  Le, Wu et~al.}]{huang2019gpipe}
\bibinfo{author}{Huang, Y.}, \bibinfo{author}{Cheng, Y.},
  \bibinfo{author}{Bapna, A.}, \bibinfo{author}{Firat, O.},
  \bibinfo{author}{Chen, D.}, \bibinfo{author}{Chen, M.}, \bibinfo{author}{Lee,
  H.}, \bibinfo{author}{Ngiam, J.}, \bibinfo{author}{Le, Q.~V.},
  \bibinfo{author}{Wu, Y.} et~al. (\bibinfo{year}{2019}).
\newblock \bibinfo{title}{Gpipe: Efficient training of giant neural networks
  using pipeline parallelism}.
\newblock {\it \bibinfo{journal}{Advances in neural information processing
  systems}\/},  {\it \bibinfo{volume}{32}\/}, \bibinfo{pages}{103--112}.
\bibitem[{Hufkens et~al.(2019)Hufkens, Melaas, Mann, Foster, Ceballos, Robles
  \& Kramer}]{hufkens2019monitoring}
\bibinfo{author}{Hufkens, K.}, \bibinfo{author}{Melaas, E.~K.},
  \bibinfo{author}{Mann, M.~L.}, \bibinfo{author}{Foster, T.},
  \bibinfo{author}{Ceballos, F.}, \bibinfo{author}{Robles, M.}, \&
  \bibinfo{author}{Kramer, B.} (\bibinfo{year}{2019}).
\newblock \bibinfo{title}{Monitoring crop phenology using a smartphone based
  near-surface remote sensing approach}.
\newblock {\it \bibinfo{journal}{Agricultural and Forest Meteorology}\/},  {\it
  \bibinfo{volume}{265}\/}, \bibinfo{pages}{327--337}.
\bibitem[{Iandola et~al.(2016)Iandola, Han, Moskewicz, Ashraf, Dally \&
  Keutzer}]{iandola2016squeezenet}
\bibinfo{author}{Iandola, F.~N.}, \bibinfo{author}{Han, S.},
  \bibinfo{author}{Moskewicz, M.~W.}, \bibinfo{author}{Ashraf, K.},
  \bibinfo{author}{Dally, W.~J.}, \& \bibinfo{author}{Keutzer, K.}
  (\bibinfo{year}{2016}).
\newblock \bibinfo{title}{Squeezenet: Alexnet-level accuracy with 50x fewer
  parameters and< 0.5 mb model size}.
\newblock {\it \bibinfo{journal}{arXiv preprint arXiv:1602.07360}\/}, .
\bibitem[{Kamilaris \& Prenafeta-Bold{\'u}(2018)}]{kamilaris2018deep}
\bibinfo{author}{Kamilaris, A.}, \& \bibinfo{author}{Prenafeta-Bold{\'u},
  F.~X.} (\bibinfo{year}{2018}).
\newblock \bibinfo{title}{Deep learning in agriculture: A survey}.
\newblock {\it \bibinfo{journal}{Computers and Electronics in Agriculture}\/},
  {\it \bibinfo{volume}{147}\/}, \bibinfo{pages}{70--90}.
\bibitem[{Keenan et~al.(2014)Keenan, Gray, Friedl, Toomey, Bohrer, Hollinger,
  Munger, O’Keefe, Schmid, Wing et~al.}]{keenan2014net}
\bibinfo{author}{Keenan, T.~F.}, \bibinfo{author}{Gray, J.},
  \bibinfo{author}{Friedl, M.~A.}, \bibinfo{author}{Toomey, M.},
  \bibinfo{author}{Bohrer, G.}, \bibinfo{author}{Hollinger, D.~Y.},
  \bibinfo{author}{Munger, J.~W.}, \bibinfo{author}{O’Keefe, J.},
  \bibinfo{author}{Schmid, H.~P.}, \bibinfo{author}{Wing, I.~S.} et~al.
  (\bibinfo{year}{2014}).
\newblock \bibinfo{title}{Net carbon uptake has increased through
  warming-induced changes in temperate forest phenology}.
\newblock {\it \bibinfo{journal}{Nature Climate Change}\/},  {\it
  \bibinfo{volume}{4}\/}, \bibinfo{pages}{598--604}.
\bibitem[{Klosterman et~al.(2014)Klosterman, Hufkens, Gray, Melaas, Sonnentag,
  Lavine, Mitchell, Norman, Friedl \& Richardson}]{klosterman2014evaluating}
\bibinfo{author}{Klosterman, S.}, \bibinfo{author}{Hufkens, K.},
  \bibinfo{author}{Gray, J.}, \bibinfo{author}{Melaas, E.},
  \bibinfo{author}{Sonnentag, O.}, \bibinfo{author}{Lavine, I.},
  \bibinfo{author}{Mitchell, L.}, \bibinfo{author}{Norman, R.},
  \bibinfo{author}{Friedl, M.}, \& \bibinfo{author}{Richardson, A.}
  (\bibinfo{year}{2014}).
\newblock \bibinfo{title}{Evaluating remote sensing of deciduous forest
  phenology at multiple spatial scales using phenocam imagery}.
\newblock {\it \bibinfo{journal}{Biogeosciences}\/},  {\it
  \bibinfo{volume}{11}\/}, \bibinfo{pages}{4305--4320}.
\bibitem[{Kussul et~al.(2014)Kussul, Skakun, Shelestov \&
  Kussul}]{kussul2014use}
\bibinfo{author}{Kussul, N.}, \bibinfo{author}{Skakun, S.},
  \bibinfo{author}{Shelestov, A.}, \& \bibinfo{author}{Kussul, O.}
  (\bibinfo{year}{2014}).
\newblock \bibinfo{title}{The use of satellite sar imagery to crop
  classification in ukraine within jecam project}.
\newblock In {\it \bibinfo{booktitle}{2014 IEEE Geoscience and Remote Sensing
  Symposium}\/} (pp. \bibinfo{pages}{1497--1500}).
\newblock \bibinfo{organization}{IEEE}.
\bibitem[{LeCun et~al.(2015{\natexlab{a}})LeCun, Bengio \&
  Hinton}]{lecun2015deep}
\bibinfo{author}{LeCun, Y.}, \bibinfo{author}{Bengio, Y.}, \&
  \bibinfo{author}{Hinton, G.} (\bibinfo{year}{2015}{\natexlab{a}}).
\newblock \bibinfo{title}{Deep learning}.
\newblock {\it \bibinfo{journal}{nature}\/},  {\it \bibinfo{volume}{521}\/},
  \bibinfo{pages}{436}.
\bibitem[{LeCun et~al.(1995)LeCun, Bengio et~al.}]{lecun1995convolutional}
\bibinfo{author}{LeCun, Y.}, \bibinfo{author}{Bengio, Y.} et~al.
  (\bibinfo{year}{1995}).
\newblock \bibinfo{title}{Convolutional networks for images, speech, and time
  series}.
\newblock {\it \bibinfo{journal}{The handbook of brain theory and neural
  networks}\/},  {\it \bibinfo{volume}{3361}\/}, \bibinfo{pages}{1995}.
\bibitem[{LeCun et~al.(2015{\natexlab{b}})}]{lecun2015lenet}
\bibinfo{author}{LeCun, Y.} et~al. (\bibinfo{year}{2015}{\natexlab{b}}).
\newblock \bibinfo{title}{Lenet-5, convolutional neural networks}.
\newblock {\it \bibinfo{journal}{URL: http://yann. lecun. com/exdb/lenet}\/},
  (p.~\bibinfo{pages}{20}).
\bibitem[{Lemoine et~al.(2013)Lemoine, Corbane, Louvrier \&
  Kauffmann}]{lemoine2013intercomparison}
\bibinfo{author}{Lemoine, G.}, \bibinfo{author}{Corbane, C.},
  \bibinfo{author}{Louvrier, C.}, \& \bibinfo{author}{Kauffmann, M.}
  (\bibinfo{year}{2013}).
\newblock \bibinfo{title}{Intercomparison and validation of building damage
  assessments based on post-haiti 2010 earthquake imagery using multi-source
  reference data}.
\newblock {\it \bibinfo{journal}{Natural Hazards and Earth System Sciences
  Discussions}\/},  {\it \bibinfo{volume}{1}\/}, \bibinfo{pages}{1445--1486}.
\bibitem[{Li \& Hoiem(2017)}]{li2017learning}
\bibinfo{author}{Li, Z.}, \& \bibinfo{author}{Hoiem, D.}
  (\bibinfo{year}{2017}).
\newblock \bibinfo{title}{Learning without forgetting}.
\newblock {\it \bibinfo{journal}{IEEE transactions on pattern analysis and
  machine intelligence}\/},  {\it \bibinfo{volume}{40}\/},
  \bibinfo{pages}{2935--2947}.
\bibitem[{Meier(1997)}]{meier1997growth}
\bibinfo{author}{Meier, U.} (\bibinfo{year}{1997}).
\newblock {\it \bibinfo{title}{Growth stages of mono-and dicotyledonous
  plants}\/}.
\newblock \bibinfo{publisher}{Blackwell Wissenschafts-Verlag}.
\bibitem[{Mohanty et~al.(2016)Mohanty, Hughes \&
  Salath{\'e}}]{mohanty2016using}
\bibinfo{author}{Mohanty, S.~P.}, \bibinfo{author}{Hughes, D.~P.}, \&
  \bibinfo{author}{Salath{\'e}, M.} (\bibinfo{year}{2016}).
\newblock \bibinfo{title}{Using deep learning for image-based plant disease
  detection}.
\newblock {\it \bibinfo{journal}{Frontiers in plant science}\/},  {\it
  \bibinfo{volume}{7}\/}, \bibinfo{pages}{1419}.
\bibitem[{Namin et~al.(2018)Namin, Esmaeilzadeh, Najafi, Brown \&
  Borevitz}]{namin2018deep}
\bibinfo{author}{Namin, S.~T.}, \bibinfo{author}{Esmaeilzadeh, M.},
  \bibinfo{author}{Najafi, M.}, \bibinfo{author}{Brown, T.~B.}, \&
  \bibinfo{author}{Borevitz, J.~O.} (\bibinfo{year}{2018}).
\newblock \bibinfo{title}{Deep phenotyping: deep learning for temporal
  phenotype/genotype classification}.
\newblock {\it \bibinfo{journal}{Plant methods}\/},  {\it
  \bibinfo{volume}{14}\/}, \bibinfo{pages}{66}.
\bibitem[{{Nationaal Georegister - BRP}(2018)}]{BRP}
\bibinfo{author}{{Nationaal Georegister - BRP}} (\bibinfo{year}{2018}).
\newblock \bibinfo{title}{Basisregistratie gewaspercelen (brp)}.
\newblock
  \bibinfo{howpublished}{\url{https://data.overheid.nl/community/dataverzoeken/basisregistratie-gewaspercelen-brp}}.
\newblock \bibinfo{note}{(Accessed on 12/08/2019)}.
\bibitem[{Nijland et~al.(2016)Nijland, Bolton, Coops \&
  Stenhouse}]{nijland2016imaging}
\bibinfo{author}{Nijland, W.}, \bibinfo{author}{Bolton, D.~K.},
  \bibinfo{author}{Coops, N.~C.}, \& \bibinfo{author}{Stenhouse, G.}
  (\bibinfo{year}{2016}).
\newblock \bibinfo{title}{Imaging phenology; scaling from camera plots to
  landscapes}.
\newblock {\it \bibinfo{journal}{Remote Sensing of Environment}\/},  {\it
  \bibinfo{volume}{177}\/}, \bibinfo{pages}{13--20}.
\bibitem[{Olivas et~al.(2009)Olivas, Guerrero, Martinez-Sober,
  Magdalena-Benedito, Serrano et~al.}]{olivas2009handbook}
\bibinfo{author}{Olivas, E.~S.}, \bibinfo{author}{Guerrero, J. D.~M.},
  \bibinfo{author}{Martinez-Sober, M.}, \bibinfo{author}{Magdalena-Benedito,
  J.~R.}, \bibinfo{author}{Serrano, L.} et~al. (\bibinfo{year}{2009}).
\newblock {\it \bibinfo{title}{Handbook of research on machine learning
  applications and trends: Algorithms, methods, and techniques: Algorithms,
  methods, and techniques}\/}.
\newblock \bibinfo{publisher}{IGI Global}.
\bibitem[{Opitz \& Burst(2019)}]{opitz2019macro}
\bibinfo{author}{Opitz, J.}, \& \bibinfo{author}{Burst, S.}
  (\bibinfo{year}{2019}).
\newblock \bibinfo{title}{Macro f1 and macro f1}.
\newblock {\it \bibinfo{journal}{arXiv preprint arXiv:1911.03347}\/}, .
\bibitem[{Paliyam et~al.(2021)Paliyam, Nakalembe, Liu, Nyiawung \&
  Kerner}]{paliyam2021street2sat}
\bibinfo{author}{Paliyam, M.}, \bibinfo{author}{Nakalembe, C.},
  \bibinfo{author}{Liu, K.}, \bibinfo{author}{Nyiawung, R.}, \&
  \bibinfo{author}{Kerner, H.} (\bibinfo{year}{2021}).
\newblock \bibinfo{title}{Street2sat: A machine learning pipeline for
  generating ground-truth geo-referenced labeled datasets from street-level
  images}.
\newblock In {\it \bibinfo{booktitle}{Tackling Climate Change with Machine
  Learning Workshop at the International Conference on Machine Learning}\/}.
\bibitem[{Pan \& Yang(2009)}]{pan2009survey}
\bibinfo{author}{Pan, S.~J.}, \& \bibinfo{author}{Yang, Q.}
  (\bibinfo{year}{2009}).
\newblock \bibinfo{title}{A survey on transfer learning}.
\newblock {\it \bibinfo{journal}{IEEE Transactions on knowledge and data
  engineering}\/},  {\it \bibinfo{volume}{22}\/}, \bibinfo{pages}{1345--1359}.
\bibitem[{Pan et~al.(2010)Pan, Yang et~al.}]{pan2010survey}
\bibinfo{author}{Pan, S.~J.}, \bibinfo{author}{Yang, Q.} et~al.
  (\bibinfo{year}{2010}).
\newblock \bibinfo{title}{A survey on transfer learning}.
\newblock {\it \bibinfo{journal}{IEEE Transactions on knowledge and data
  engineering}\/},  {\it \bibinfo{volume}{22}\/}, \bibinfo{pages}{1345--1359}.
\bibitem[{Ringland et~al.(2019)Ringland, Bohm \&
  Baek}]{ringland2019characterization}
\bibinfo{author}{Ringland, J.}, \bibinfo{author}{Bohm, M.}, \&
  \bibinfo{author}{Baek, S.-R.} (\bibinfo{year}{2019}).
\newblock \bibinfo{title}{Characterization of food cultivation along roadside
  transects with google street view imagery and deep learning}.
\newblock {\it \bibinfo{journal}{Computers and electronics in agriculture}\/},
  {\it \bibinfo{volume}{158}\/}, \bibinfo{pages}{36--50}.
\bibitem[{Sabottke \& Spieler(2020)}]{sabottke2020effect}
\bibinfo{author}{Sabottke, C.~F.}, \& \bibinfo{author}{Spieler, B.~M.}
  (\bibinfo{year}{2020}).
\newblock \bibinfo{title}{The effect of image resolution on deep learning in
  radiography}.
\newblock {\it \bibinfo{journal}{Radiology: Artificial Intelligence}\/},  {\it
  \bibinfo{volume}{2}\/}, \bibinfo{pages}{e190015}.
\bibitem[{Sandler et~al.(2018)Sandler, Howard, Zhu, Zhmoginov \&
  Chen}]{sandler2018mobilenetv2}
\bibinfo{author}{Sandler, M.}, \bibinfo{author}{Howard, A.},
  \bibinfo{author}{Zhu, M.}, \bibinfo{author}{Zhmoginov, A.}, \&
  \bibinfo{author}{Chen, L.-C.} (\bibinfo{year}{2018}).
\newblock \bibinfo{title}{Mobilenetv2: Inverted residuals and linear
  bottlenecks}.
\newblock In {\it \bibinfo{booktitle}{Proceedings of the IEEE Conference on
  Computer Vision and Pattern Recognition}\/} (pp.
  \bibinfo{pages}{4510--4520}).
\bibitem[{Schaap et~al.(2011)Schaap, Reidsma, Mandryk, Verhagen, van~der Wal,
  Wolf \& Van~Ittersum}]{schaap2011adapting}
\bibinfo{author}{Schaap, B.~F.}, \bibinfo{author}{Reidsma, P.},
  \bibinfo{author}{Mandryk, M.}, \bibinfo{author}{Verhagen, A.},
  \bibinfo{author}{van~der Wal, M.}, \bibinfo{author}{Wolf, J.}, \&
  \bibinfo{author}{Van~Ittersum, M.} (\bibinfo{year}{2011}).
\newblock {\it \bibinfo{title}{Adapting agriculture in 2050 in Flevoland;
  perspectives from stakeholders}\/}.
\newblock \bibinfo{type}{Technical Report} Wageningen UR.
\bibitem[{Seiferling et~al.(2017)Seiferling, Naik, Ratti \&
  Proulx}]{seiferling2017green}
\bibinfo{author}{Seiferling, I.}, \bibinfo{author}{Naik, N.},
  \bibinfo{author}{Ratti, C.}, \& \bibinfo{author}{Proulx, R.}
  (\bibinfo{year}{2017}).
\newblock \bibinfo{title}{Green streets- quantifying and mapping urban trees
  with street-level imagery and computer vision}.
\newblock {\it \bibinfo{journal}{Landscape and Urban Planning}\/},  {\it
  \bibinfo{volume}{165}\/}, \bibinfo{pages}{93--101}.
\bibitem[{Soille et~al.(2018)Soille, Burger, De~Marchi, Kempeneers, Rodriguez,
  Syrris \& Vasilev}]{soille2018versatile}
\bibinfo{author}{Soille, P.}, \bibinfo{author}{Burger, A.},
  \bibinfo{author}{De~Marchi, D.}, \bibinfo{author}{Kempeneers, P.},
  \bibinfo{author}{Rodriguez, D.}, \bibinfo{author}{Syrris, V.}, \&
  \bibinfo{author}{Vasilev, V.} (\bibinfo{year}{2018}).
\newblock \bibinfo{title}{A versatile data-intensive computing platform for
  information retrieval from big geospatial data}.
\newblock {\it \bibinfo{journal}{Future Generation Computer Systems}\/},  {\it
  \bibinfo{volume}{81}\/}, \bibinfo{pages}{30--40}.
\bibitem[{Sonnentag et~al.(2012)Sonnentag, Hufkens, Teshera-Sterne, Young,
  Friedl, Braswell, Milliman, O’Keefe \& Richardson}]{sonnentag2012digital}
\bibinfo{author}{Sonnentag, O.}, \bibinfo{author}{Hufkens, K.},
  \bibinfo{author}{Teshera-Sterne, C.}, \bibinfo{author}{Young, A.~M.},
  \bibinfo{author}{Friedl, M.}, \bibinfo{author}{Braswell, B.~H.},
  \bibinfo{author}{Milliman, T.}, \bibinfo{author}{O’Keefe, J.}, \&
  \bibinfo{author}{Richardson, A.~D.} (\bibinfo{year}{2012}).
\newblock \bibinfo{title}{Digital repeat photography for phenological research
  in forest ecosystems}.
\newblock {\it \bibinfo{journal}{Agricultural and Forest Meteorology}\/},  {\it
  \bibinfo{volume}{152}\/}, \bibinfo{pages}{159--177}.
\bibitem[{Stafford et~al.(2010)Stafford, Hart, Collins, Kirkhope, Williams,
  Rees, Lloyd \& Goodenough}]{stafford2010eu}
\bibinfo{author}{Stafford, R.}, \bibinfo{author}{Hart, A.~G.},
  \bibinfo{author}{Collins, L.}, \bibinfo{author}{Kirkhope, C.~L.},
  \bibinfo{author}{Williams, R.~L.}, \bibinfo{author}{Rees, S.~G.},
  \bibinfo{author}{Lloyd, J.~R.}, \& \bibinfo{author}{Goodenough, A.~E.}
  (\bibinfo{year}{2010}).
\newblock \bibinfo{title}{Eu-social science: the role of internet social
  networks in the collection of bee biodiversity data}.
\newblock {\it \bibinfo{journal}{PloS one}\/},  {\it \bibinfo{volume}{5}\/},
  \bibinfo{pages}{e14381}.
\bibitem[{Sun et~al.(2009)Sun, Wong \& Kamel}]{sun2009classification}
\bibinfo{author}{Sun, Y.}, \bibinfo{author}{Wong, A.~K.}, \&
  \bibinfo{author}{Kamel, M.~S.} (\bibinfo{year}{2009}).
\newblock \bibinfo{title}{Classification of imbalanced data: A review}.
\newblock {\it \bibinfo{journal}{International journal of pattern recognition
  and artificial intelligence}\/},  {\it \bibinfo{volume}{23}\/},
  \bibinfo{pages}{687--719}.
\bibitem[{Tan \& Le(2019)}]{tan2019efficientnet}
\bibinfo{author}{Tan, M.}, \& \bibinfo{author}{Le, Q.} (\bibinfo{year}{2019}).
\newblock \bibinfo{title}{Efficientnet: Rethinking model scaling for
  convolutional neural networks}.
\newblock In {\it \bibinfo{booktitle}{International Conference on Machine
  Learning}\/} (pp. \bibinfo{pages}{6105--6114}).
\newblock \bibinfo{organization}{PMLR}.
\bibitem[{Waldner et~al.(2019)Waldner, Bellemans, Hochman, Newby, de~Abelleyra,
  Ver{\'o}n, Bartalev, Lavreniuk, Kussul, Le~Maire
  et~al.}]{waldner2019roadside}
\bibinfo{author}{Waldner, F.}, \bibinfo{author}{Bellemans, N.},
  \bibinfo{author}{Hochman, Z.}, \bibinfo{author}{Newby, T.},
  \bibinfo{author}{de~Abelleyra, D.}, \bibinfo{author}{Ver{\'o}n, S.~R.},
  \bibinfo{author}{Bartalev, S.}, \bibinfo{author}{Lavreniuk, M.},
  \bibinfo{author}{Kussul, N.}, \bibinfo{author}{Le~Maire, G.} et~al.
  (\bibinfo{year}{2019}).
\newblock \bibinfo{title}{Roadside collection of training data for cropland
  mapping is viable when environmental and management gradients are surveyed}.
\newblock {\it \bibinfo{journal}{International Journal of Applied Earth
  Observation and Geoinformation}\/},  {\it \bibinfo{volume}{80}\/},
  \bibinfo{pages}{82--93}.
\bibitem[{Wu et~al.(2021)Wu, Wu, Zhang, Zeng \& Tian}]{wu2021identification}
\bibinfo{author}{Wu, F.}, \bibinfo{author}{Wu, B.}, \bibinfo{author}{Zhang,
  M.}, \bibinfo{author}{Zeng, H.}, \& \bibinfo{author}{Tian, F.}
  (\bibinfo{year}{2021}).
\newblock \bibinfo{title}{Identification of crop type in crowdsourced road view
  photos with deep convolutional neural network}.
\newblock {\it \bibinfo{journal}{Sensors}\/},  {\it \bibinfo{volume}{21}\/},
  \bibinfo{pages}{1165}.
\bibitem[{Wu et~al.(2019)Wu, Kirillov, Massa, Lo \&
  Girshick}]{wu2019detectron2}
\bibinfo{author}{Wu, Y.}, \bibinfo{author}{Kirillov, A.},
  \bibinfo{author}{Massa, F.}, \bibinfo{author}{Lo, W.-Y.}, \&
  \bibinfo{author}{Girshick, R.} (\bibinfo{year}{2019}).
\newblock \bibinfo{title}{Detectron2}.
\newblock
  \bibinfo{howpublished}{\url{https://github.com/facebookresearch/detectron2}}.
\bibitem[{Yalcin(2015)}]{yalcin2015phenology}
\bibinfo{author}{Yalcin, H.} (\bibinfo{year}{2015}).
\newblock \bibinfo{title}{Phenology monitoring of agricultural plants using
  texture analysis}.
\newblock In {\it \bibinfo{booktitle}{Agro-Geoinformatics
  (Agro-geoinformatics), 2015 Fourth International Conference on}\/} (pp.
  \bibinfo{pages}{338--342}).
\newblock \bibinfo{organization}{IEEE}.
\bibitem[{Yalcin(2017)}]{yalcin2017plant}
\bibinfo{author}{Yalcin, H.} (\bibinfo{year}{2017}).
\newblock \bibinfo{title}{Plant phenology recognition using deep learning:
  Deep-pheno}.
\newblock In {\it \bibinfo{booktitle}{Agro-Geoinformatics, 2017 6th
  International Conference on}\/} (pp. \bibinfo{pages}{1--5}).
\newblock \bibinfo{organization}{IEEE}.
\bibitem[{Yan \& Ryu(2021)}]{yan2021exploring}
\bibinfo{author}{Yan, Y.}, \& \bibinfo{author}{Ryu, Y.} (\bibinfo{year}{2021}).
\newblock \bibinfo{title}{Exploring google street view with deep learning for
  crop type mapping}.
\newblock {\it \bibinfo{journal}{ISPRS Journal of Photogrammetry and Remote
  Sensing}\/},  {\it \bibinfo{volume}{171}\/}, \bibinfo{pages}{278--296}.
\bibitem[{Zhang et~al.(2018)Zhang, Zhou, Lin \& Sun}]{zhang2018shufflenet}
\bibinfo{author}{Zhang, X.}, \bibinfo{author}{Zhou, X.}, \bibinfo{author}{Lin,
  M.}, \& \bibinfo{author}{Sun, J.} (\bibinfo{year}{2018}).
\newblock \bibinfo{title}{Shufflenet: An extremely efficient convolutional
  neural network for mobile devices}.
\newblock In {\it \bibinfo{booktitle}{Proceedings of the IEEE conference on
  computer vision and pattern recognition}\/} (pp.
  \bibinfo{pages}{6848--6856}).
\bibitem[{Zheng et~al.(2019)Zheng, Kong, Jin, Wang, Su \&
  Zuo}]{zheng2019cropdeep}
\bibinfo{author}{Zheng, Y.-Y.}, \bibinfo{author}{Kong, J.-L.},
  \bibinfo{author}{Jin, X.-B.}, \bibinfo{author}{Wang, X.-Y.},
  \bibinfo{author}{Su, T.-L.}, \& \bibinfo{author}{Zuo, M.}
  (\bibinfo{year}{2019}).
\newblock \bibinfo{title}{Cropdeep: The crop vision dataset for
  deep-learning-based classification and detection in precision agriculture}.
\newblock {\it \bibinfo{journal}{Sensors}\/},  {\it \bibinfo{volume}{19}\/},
  \bibinfo{pages}{1058}.
\bibitem[{Zoph et~al.(2018)Zoph, Vasudevan, Shlens \& Le}]{zoph2018learning}
\bibinfo{author}{Zoph, B.}, \bibinfo{author}{Vasudevan, V.},
  \bibinfo{author}{Shlens, J.}, \& \bibinfo{author}{Le, Q.~V.}
  (\bibinfo{year}{2018}).
\newblock \bibinfo{title}{Learning transferable architectures for scalable
  image recognition}.
\newblock In {\it \bibinfo{booktitle}{Proceedings of the IEEE conference on
  computer vision and pattern recognition}\/} (pp.
  \bibinfo{pages}{8697--8710}).

\end{thebibliography}
